\documentclass[11pt]{article}

\usepackage[final]{acl}

\usepackage{times}
\usepackage{latexsym}
\usepackage{amsmath}
\usepackage{amssymb}
\usepackage{booktabs}
\usepackage{longtable}
\usepackage{multirow}
\usepackage{amsfonts}
\usepackage{xltabular}
\usepackage{tabularx}
\usepackage{capt-of}
\usepackage{makecell}
\usepackage[table]{xcolor}
\usepackage{graphicx}
\usepackage[T1]{fontenc}

\usepackage[utf8]{inputenc}

\usepackage{microtype}

\usepackage{inconsolata}

\usepackage{graphicx}

\usepackage{colortbl}

\title{Behavior-Preserving KV Cache Compression}

\author{
  \textbf{Doo Hwan Hwang}\textsuperscript{1},
  \textbf{Junyoung Jang}\textsuperscript{1},
  \textbf{Junho Na}\textsuperscript{1},
  \textbf{Hosung Lim}\textsuperscript{2},
  \textbf{Kee-Eung Kim}\textsuperscript{1}
  \\
  \textsuperscript{1}Kim Jaechul Graduate School of AI, KAIST,
  Daejeon, Republic of Korea
  \\
  \textsuperscript{2}KT Corporation, Seoul, Republic of Korea
  \\
  \small{
    \textbf{Correspondence:}
    \href{mailto:dhhwang@ai.kaist.ac.kr}
    {dhhwang@ai.kaist.ac.kr}
  }
}

\begin{document}
\maketitle
\begin{abstract}
KV caches are a major bottleneck in long-context inference and long-form generation with large language models.
Existing training-free eviction policies largely rely on proxy importance signals, such as attention mass, to decide which past tokens to retain. 
We argue that cache compression should instead preserve the predictive behavior
of the full-cache model, retaining entries whose removal would substantially
change the model's output distribution.
We propose Behavior-Preserving KV Cache Compression, a training-free framework that scores candidate evictions by estimating the compressed-cache logits induced by their removal and evaluating the resulting KL to the full-cache next-token distribution. Using pre-eviction forward statistics, the method avoids running separate masked forward passes for each candidate.
Across diverse architectures and both prefill-time and generation-time compression, our method delivers substantial gains in downstream task quality over lightweight attention-based heuristics at matched retained-KV budgets, with the largest gains under aggressive compression.
It achieves these gains with additional compression-time computation while retaining an end-to-end speedup over full-cache inference in our evaluated settings.
\end{abstract}

\section{Introduction}
Large language models (LLMs) rely on key-value (KV) caches during autoregressive
decoding to avoid recomputing past activations
\cite{DBLP:conf/nips/Zhang00CZC0TRBW23,DBLP:journals/corr/abs-2510-00636}.
As sequence lengths grow, these caches scale linearly with the number of
processed tokens and become a major bottleneck for memory usage and attention
computation
\cite{DBLP:conf/nips/LiuDLWXXKS23,DBLP:conf/nips/LiHYVLYCLC24,DBLP:conf/nips/Zhang00CZC0TRBW23}.
KV cache compression reduces this cost by retaining only a subset of cached
key-value entries. We focus on training-free cache management, where the
eviction rule is computed from inference-time statistics rather than learned
from task-specific traces or optimized through additional cache-policy training.
This setting is attractive for deployment because it can be applied to
off-the-shelf LLMs across different budgets, context lengths, and tasks without
training an auxiliary eviction policy.

In this setting, the natural reference is the behavior of the same model before
eviction. A compressed cache should therefore not merely satisfy a memory budget;
it should preserve the predictive behavior of the pre-eviction computation.
Since future decoding states are unavailable when compression decisions are made,
we use the next-token distribution at available query positions as an observable
reference behavior.

Existing training-free methods approximate this goal through indirect proxy
signals. Attention-based methods retain entries that receive high attention in
the current context
\citep{DBLP:conf/nips/LiuDLWXXKS23,DBLP:conf/nips/LiHYVLYCLC24,
DBLP:conf/nips/Zhang00CZC0TRBW23} or are expected to receive high attention
during future decoding~\citep{DBLP:journals/corr/abs-2510-00636}. 
Other methods favor entries that help reconstruct the original context from the compressed cache~\cite{kim2026kvzip}. These proxy objectives need not match the predictive effect of eviction: attention can overvalue redundant entries, while reconstruction utility can favor context recovery rather than preserving the next-token distribution. This mismatch motivates scoring cache entries by their estimated effect on model predictions.

We propose a \textbf{behavior-preserving framework} for KV cache compression. Our method
scores candidate evictions by estimating their effect on the full-cache
next-token distribution. It estimates the compressed-cache logits induced by
eviction from pre-eviction forward statistics using a residual-path
approximation, and plugs the estimated logits into a KL objective against the
full-cache prediction. This avoids candidate-specific masked forward passes
while ranking entries by their predicted behavioral effect.

We evaluate the method in two inference settings. In prefill-time compression,
the prompt cache is compressed after long-context encoding. In generation-time
compression, generated-token caches are compressed online during autoregressive
decoding. Across models, datasets, and cache budgets, our method delivers substantially higher downstream task quality than lightweight attention-based eviction methods at matched retained-KV budgets, particularly under aggressive compression.
Its behavior-preserving scoring incurs additional compression-time computation, while the resulting compressed inference remains faster than full-cache inference in our evaluated settings.

Our contributions are as follows:
\begin{itemize}
    \item We recast training-free KV cache compression as behavior-preserving
    eviction, evaluating cached entries by their predicted effect on the full-cache
    next-token distribution rather than by proxy signals such as attention mass or
    reconstruction utility.

    \item We introduce a forward-only scoring method that estimates
    compressed-cache logits from pre-eviction forward statistics and evaluates
    their KL to the full-cache prediction, without candidate-specific masked
    forward passes.

    \item We provide a broad evaluation in both prefill-time and generation-time compression settings, showing that behavior-preserving scoring yields stronger compression--performance trade-offs, especially under aggressive cache budgets.
\end{itemize}

\section{Related Work}
KV cache compression reduces the memory and attention cost of long-context
inference by reducing the stored KV states. Learned KV cache compression methods
modify the model or train an auxiliary mechanism to improve the
compression--quality trade-off. Dynamic Memory Compression learns head- and
layer-specific compression ratios through continued pretraining
\citep{nawrot2024dynamicmemorycompression}, while Dynamic Memory Sparsification
trains a sparse cache policy for inference-time scaling
\citep{lancucki2025inferencetimehyperscaling}. These approaches can be
effective, but they require additional training or model retrofitting.

Training-free KV cache compression instead computes compression decisions from
inference-time statistics. Early token-selection methods exploit attention
persistence by retaining heavy hitters, recent tokens, or tokens that received
high attention scores in previous decoding steps
\citep{DBLP:conf/nips/LiuDLWXXKS23, DBLP:conf/nips/Zhang00CZC0TRBW23}.
SnapKV extends this idea to prompt compression by selecting entries from
attention patterns observed near the end of the prompt, while PyramidKV combines
attention-based selection with layer-wise cache budgets
\citep{DBLP:conf/nips/LiHYVLYCLC24, DBLP:journals/corr/abs-2406-02069}.
Orthogonal to token selection, KIVI compresses the KV cache by reducing numerical
precision through tuning-free quantization \citep{liu2024kivi}.

Recent training-free KV cache compression methods use richer proxy objectives.
Expected Attention estimates future attention from a closed-form model of
future-query distributions \citep{DBLP:journals/corr/abs-2510-00636}. KVzip
scores cached entries by their contribution to reconstructing the original
context, enabling query-agnostic compressed caches \citep{kim2026kvzip}. CAOTE
estimates the attention-output error caused by token removal
\citep{goel2026caote}, which captures an eviction-induced perturbation in
attention-output space but not its resulting effect on the model's output
distribution. R-KV combines token importance with redundancy for
reasoning-model decoding \citep{cai2026rkv}. 
These methods broaden KV cache compression beyond raw attention scores, but they still optimize proxy objectives such as future attention, reconstruction utility, attention-output error, or redundancy. In contrast, our method scores evictions by their estimated effect on the full-cache next-token distribution, aligning cache retention with predictive-behavior preservation rather than proxy importance.



\begin{figure*}[t]
    \centering
    \includegraphics[width=\linewidth]{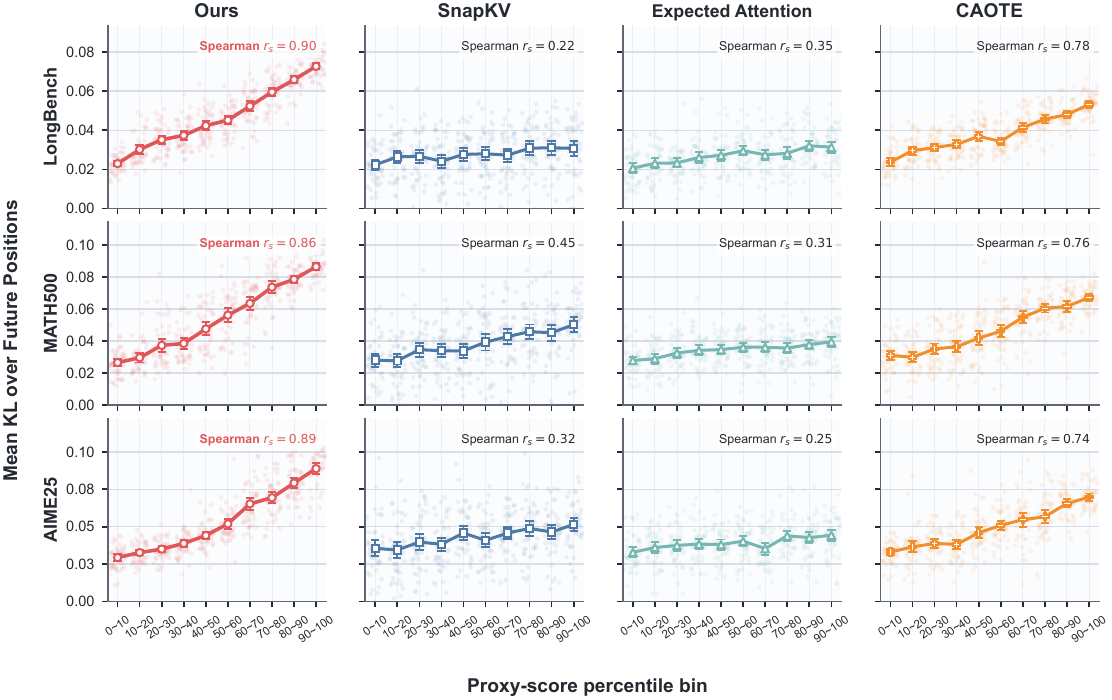}
\caption{
Relationship between method-specific proxy scores and KL over future
positions on LongBench, MATH500, and AIME25 using LLaMA3.1-8B. For each scoring
rule, we group candidate eviction masks by the percentile of the method's own
score and sample perturbed KV caches from each score bin. We then measure the
mean KL over future positions along the same reference continuation. The x-axis
reports the score percentile. Higher percentiles should correspond to larger KL
over future positions if a score captures the predictive effect of eviction.
The behavior-preserving score exhibits the clearest association, while the
association is less pronounced for proxy-based scores.
}
\label{fig:next_token_kl_quantile}
\end{figure*}

\section{Motivation and Analysis}
\subsection{Behavior Preservation at Available Query Positions}
\label{sec:why_next_token}

KV cache compression aims to reduce memory and attention cost while preserving
the predictive behavior of the full-cache computation. Because a decoder-only
language model defines a distribution over continuations through a product of
next-token distributions, preserving behavior over a future continuation can be
stated as preserving the autoregressive distribution induced by the compressed
cache.

Let \(y_{\le t}\) denote the prefix observed up to position \(t\). For a future
continuation length \(\tau\), let \(P_t^\tau\) and \(Q_t^\tau\) denote the
conditional distributions over \(y_{t+1:t+\tau}\) induced by the full and
compressed KV caches, respectively. A natural behavior-preserving objective is
\[
D_{\mathrm{KL}}\!\left(
P_t^\tau
\,\middle\|\,
Q_t^\tau
\right).
\]
Let \(p_t\) and \(q_t\) be the corresponding next-token distributions at the
current prefix. By the chain rule for autoregressive KL,
\[
\begin{aligned}
D_{\mathrm{KL}}\!\left(P_t^\tau \,\middle\|\, Q_t^\tau\right)
&=
D_{\mathrm{KL}}\!\left(p_t \,\middle\|\, q_t\right)
\\
&+
\mathbb{E}_{y_{t+1}\sim p_t}
\left[
D_{\mathrm{KL}}\!\left(
P_{t+1}^{\tau-1}
\,\middle\|\,
Q_{t+1}^{\tau-1}
\right)
\right].
\end{aligned}
\]

Thus, the next-token KL is the first term of the continuation-level behavioral
discrepancy. If compression substantially changes this term, then it has already
changed the model's predictive behavior at the current prefix, independently of
how later states evolve.

The remaining terms depend on future tokens and future query states, which are
not available when compression is applied. Estimating them would require
rollouts, a learned model of future queries, or task-specific assumptions. We
therefore use the observable component of the behavior-preserving objective. For
the set of query positions \(\mathcal T\) available at compression time, we
define
\[
\mathcal J
=
\frac{1}{|\mathcal T|}
\sum_{t\in\mathcal T}
D_{\mathrm{KL}}\!\left(p_t \,\middle\|\, q_t\right).
\]
This quantity measures compression error directly in output-distribution space
and can be evaluated before future queries are realized.

Next-token KL does not fully characterize future behavior, since later queries
may amplify or dampen the perturbation introduced by compression. Nevertheless,
it is the observable component of the long-horizon behavioral divergence at
compression time. Figure~\ref{fig:next_token_kl_quantile} supports using an
output-distribution-based compression score: using LLaMA3.1-8B, we group sampled
eviction masks by each method's score percentile and measure the mean KL over
subsequent positions along the same reference continuation. Higher score
percentiles should correspond to larger future-position KL if the score captures
the predictive effect of eviction. The behavior-preserving score shows the
clearest association, while the association is less pronounced for proxy-based
scores. We therefore use next-token KL as the compression-time behavioral target
and score evictions by their estimated output-distribution perturbation.

\section{Behavior-Preserving KV Cache Compression}
\subsection{Behavior-Preserving Objective}
\label{sec:method_objective}

Let the pre-eviction KV cache contain \(S\) token positions across \(L\)
Transformer layers. We use a cache-retention mask
\(d\in\{0,1\}^{L\times S}\), where \(d_{\ell j}=1\) retains token \(j\)'s KV
entry at layer \(\ell\), and \(d_{\ell j}=0\) evicts it. We also define the
corresponding eviction indicator \(\bar d_{\ell j}=1-d_{\ell j}\). The mask is
shared across KV heads within each layer.

As discussed in Section~\ref{sec:why_next_token}, we use the next-token
distribution at available query positions as the compression-time behavioral
target. Let \(\mathcal T\) denote these positions: prompt-end positions for
prefill-time compression and recent generated positions for generation-time
compression. For each \(t\in\mathcal T\), let \(z_t\) be the full-cache
vocabulary logits and \(p_t=\mathrm{softmax}(z_t)\). If a retention mask \(d\) induces the exact logit perturbation
\(\Delta z_{t,d}\), the compressed-cache prediction is
\[
q_t^{(d)}=\mathrm{softmax}(z_t+\Delta z_{t,d}).
\]
The ideal mask minimizes the KL from the full-cache prediction to the
compressed-cache prediction:
\begin{equation}
\mathcal J(d)
=
\frac{1}{|\mathcal T|}
\sum_{t\in\mathcal T}
D_{\mathrm{KL}}\!\left(p_t \,\middle\|\, q_t^{(d)}\right).
\label{eq:behavior_objective}
\end{equation}

This objective makes the target explicit, but it is not directly usable for
selection because obtaining \(\Delta z_{t,d}\) for many candidate masks would
require candidate-specific masked-cache forward passes. We instead estimate
\(\Delta z_{t,d}\) from full-cache forward statistics by first computing
layerwise perturbations and then propagating them to the logits.

\subsection{Computing Layer Perturbations from KV Eviction}
\label{sec:layerwise_perturbation}
We first estimate how a layer-level cache mask changes the attention output in
the residual stream. This gives a layerwise residual perturbation
\(\Delta u_{\ell t}(d_\ell)\), which will later be propagated to the vocabulary
logits. For layer \(\ell\), attention head \(r\), and query position
\(t\in\mathcal T\), let \(a_{\ell rtj}\) be the pre-eviction attention weight
assigned to token \(j\). Positions not visible under the causal mask have
attention weight zero. Let \(W^O_{\ell r}\) denote the block of the
layer-\(\ell\) attention output projection corresponding to head \(r\). We write
\[
\tilde v_{\ell rj}=W^O_{\ell r}v_{\ell rj},
\quad
o_{\ell rt}=\sum_{i=1}^{S}a_{\ell rti}\tilde v_{\ell ri},
\]
where \(\tilde v_{\ell rj}\) is token \(j\)'s output-projected value and
\(o_{\ell rt}\) is the full-cache projected head contribution.

For a layer mask \(d_\ell\), we use a deletion counterfactual that holds the
pre-eviction query, keys, and values fixed, removes masked entries, and
renormalizes attention over the retained entries. The resulting projected
head-output perturbation is
\begin{equation}
\Delta o_{\ell rt}(d_\ell)
=
\frac{
\sum_{i=1}^{S}d_{\ell i}a_{\ell rti}\tilde v_{\ell ri}
}{
\sum_{i=1}^{S}d_{\ell i}a_{\ell rti}
}
-
o_{\ell rt}.
\label{eq:head_perturbation}
\end{equation}
The denominator is the retained attention mass, so
Eq.~\eqref{eq:head_perturbation} is the difference between the renormalized
retained-cache head contribution and the full-cache head contribution.

Summing over heads gives the residual-stream perturbation injected by the
attention block at layer \(\ell\):
\begin{equation}
\Delta u_{\ell t}(d_\ell)
=
\sum_{r=1}^{H}
\Delta o_{\ell rt}(d_\ell).
\label{eq:layer_perturbation}
\end{equation}
Equations~\eqref{eq:head_perturbation} and
\eqref{eq:layer_perturbation} are computed from full-cache attention weights and
cached values. They define the deletion-induced residual perturbations that are
propagated to logits in the next subsection.

\subsection{From Layer Perturbations to Logit Perturbations}
\label{sec:residual_path}

The logits \(z_t\) in Section~\ref{sec:method_objective} are produced from the
final residual state of the Transformer. Let \(x_t^L\) be the full-cache final
pre-normalization residual state at position \(t\). The corresponding logits are
\[
z_t = W N_f(x_t^L) + b,
\]
where \(N_f\) is the final normalization, \(W\) is the LM head, and \(b\) is the
LM-head bias if present. If a mask \(d\) changes the final residual state by
\(\Delta x^L_{t,d}\), then a first-order expansion around the full-cache state
gives
\begin{equation}
\Delta z_{t,d}
\approx
WJ_t\Delta x^L_{t,d},
\quad
J_t = \nabla N_f(x_t^L).
\label{eq:first_order_logit}
\end{equation}

Computing \(\Delta x^L_{t,d}\) exactly would require an additional forward
computation under the masked cache. Let \(\Delta u_{\ell t}(d_\ell)\) denote
the residual-stream perturbation injected by the attention block at layer
\(\ell\) under the layer mask \(d_\ell\). Since attention outputs are added to
the residual stream, we approximate the final-state perturbation by the direct
residual-path contribution:
\begin{equation}
\widehat{\Delta x}^L_{t,d}
=
\sum_{\ell=1}^{L}
\Delta u_{\ell t}(d_\ell).
\label{eq:residual_path_approx}
\end{equation}
This approximation keeps the component computable from pre-eviction statistics
and treats the responses of later attention, MLP, and normalization layers as
approximation error. We validate this residual-path ranking signal in
Appendix~\ref{app:residual_path_validation}.

Substituting Eq.~\eqref{eq:residual_path_approx} into
Eq.~\eqref{eq:first_order_logit} gives the estimated logit perturbation
\begin{equation}
\widehat{\Delta z}_{t,d}
=
WJ_t
\sum_{\ell=1}^{L}
\Delta u_{\ell t}(d_\ell).
\label{eq:logit_surrogate}
\end{equation}
We then form the estimated compressed-cache logits
\begin{equation}
\hat z_{t,d}
=
z_t+\widehat{\Delta z}_{t,d}.
\label{eq:estimated_logits}
\end{equation}
Plugging these logits into Eq.~\eqref{eq:behavior_objective} gives the practical
behavior-preserving score:
\begin{equation}
\begin{aligned}
\widehat{\mathcal J}(d)
&=
\frac{1}{|\mathcal T|}
\sum_{t\in\mathcal T}
D_{\mathrm{KL}}\!\left(
p_t \,\middle\|\, \hat q_t^{(d)}
\right), \\
\hat q_t^{(d)}
&=
\mathrm{softmax}(\hat z_{t,d}).
\end{aligned}
\label{eq:estimated_behavior_objective}
\end{equation}
Small values of \(\widehat{\mathcal J}(d)\) indicate masks whose estimated
compressed-cache logits preserve the full-cache predictive distribution.

\subsection{Budgeted Mask Selection}
\label{sec:mask_selection}

We select \(B\) layer-token entries to evict. Let \(\mathcal P\) denote
protected entries that must be retained, and let \(\mathcal E\) denote the
eligible entries. Using the eviction indicator \(\bar d_{\ell j}=1-d_{\ell j}\),
the feasible masks are
\begin{equation*}
\begin{aligned}
\mathcal D_B
=
\bigl\{
d\in\{0,1\}^{L\times S}
\,\big|\,&
\bar d_{\ell j}=0\ \forall(\ell,j)\in\mathcal P,\\
&
\sum_{(\ell,j)\in\mathcal E}\bar d_{\ell j}=B
\bigr\}.
\end{aligned}
\end{equation*}
The score in Eq.~\eqref{eq:estimated_behavior_objective} evaluates the
behavioral effect after applying a retention mask, but directly minimizing it
over \(\mathcal D_B\) would require searching over all subsets of \(B\) evicted
entries. This is intractable because eviction
costs are non-additive: removing multiple entries changes both the renormalized
attention outputs and the resulting predictive distribution. We therefore use a
two-stage greedy procedure.

First, we form a candidate pool using singleton scores. For an eligible entry
\((\ell,j)\), we consider the singleton eviction indicated by
\(\bar d_{\ell j}=1\), while all other eligible eviction indicators remain zero.
The
corresponding single-entry logit estimate is
\[
\hat z_{t,(\ell,j)}
=
z_t
+
WJ_t
\Delta u_{\ell t}
\!\left(
\bar d_{\ell j}
\right).
\]
We score this singleton eviction by plugging the estimated logits into the
behavior-preserving KL:
\[
\frac{1}{|\mathcal T|}
\sum_{t\in\mathcal T}
D_{\mathrm{KL}}\!\left(
p_t
\,\middle\|\,
\mathrm{softmax}(\hat z_{t,(\ell,j)})
\right).
\]
We construct \(\mathcal C_{\mathrm{pool}}\) from the eligible entries with the
lowest singleton-eviction scores.

Second, we refine the selected evictions using mask-level marginal costs.
Starting from an empty selected set \(\mathcal M\), we repeatedly add the
candidate in \(\mathcal C_{\mathrm{pool}}\) that causes the smallest increase in
the estimated-logit KL score from Eq.~\eqref{eq:estimated_behavior_objective},
updating cached mask statistics after each step. We continue until \(B\) entries
have been selected. This refinement accounts for interactions among evicted
entries, including attention renormalization and changes in the estimated
compressed-cache distribution. Implementation details, including top-\(K\) KL
evaluation and refinement batching, are given in Appendix~\ref{app:implementation}.


\section{Experiments}
\definecolor{oursrow}{HTML}{EAF2FB}   
\definecolor{headrow}{HTML}{F2F2F2}   

\begin{table*}[p]
\centering
\footnotesize
\setlength{\tabcolsep}{4.8pt}
\renewcommand{\arraystretch}{0.93}
\begin{tabular}{@{}lllcccccc@{}}
\toprule
\rowcolor{headrow}
\textbf{Model} & \textbf{Metric} & \textbf{Method} & \textbf{Full}
& $\rho\!=\!0.10$ & $\rho\!=\!0.25$ & $\rho\!=\!0.50$
& $\rho\!=\!0.75$ & $\rho\!=\!0.90$ \\
\midrule

\multirow[t]{18}{*}{\textbf{Qwen3-8B}}
& \multirow{6}{*}{LongBench Avg.} & SnapKV & \multirow{6}{*}{48.9} & 47.5 & 47.0 & 45.1 & 40.7 & 32.2 \\
& & PyramidKV & & 47.4 & 46.8 & 44.7 & 40.0 & 33.0 \\
& & KVzip     & & 47.7 & 47.2 & 45.9 & 42.7 & 36.1 \\
& & EA        & & 48.0 & 47.6 & 47.2 & 45.8 & 38.3 \\
& & CAOTE     & & 47.8 & 47.6 & 47.0 & 44.4 & 37.2 \\
\rowcolor{oursrow}
& & \textbf{Ours} & & \textbf{48.2} & \textbf{48.0} & \textbf{47.9} & \textbf{47.4} & \textbf{41.6} \\
\cmidrule(lr){2-9}

& \multirow{6}{*}{RULER 8K} & SnapKV & \multirow{6}{*}{86.1} & 85.4 & 84.5 & 78.8 & 62.7 & 47.9 \\
& & PyramidKV & & 85.6 & 84.3 & 80.6 & 64.8 & 49.3 \\
& & KVzip     & & 85.7 & 85.0 & 81.8 & 66.2 & 52.4 \\
& & EA        & & 85.5 & 85.2 & 80.7 & 69.4 & 54.0 \\
& & CAOTE     & & 85.8 & 84.9 & 82.3 & 70.1 & 56.7 \\
\rowcolor{oursrow}
& & \textbf{Ours} & & \textbf{86.0} & \textbf{85.6} & \textbf{83.0} & \textbf{75.4} & \textbf{66.2} \\
\cmidrule(lr){2-9}

& \multirow{6}{*}{RULER 16K} & SnapKV & \multirow{6}{*}{79.5} & 78.7 & 77.4 & 70.8 & 54.9 & 37.1 \\
& & PyramidKV & & 79.0 & 77.7 & 72.5 & 56.4 & 39.6 \\
& & KVzip     & & 78.9 & 77.8 & 73.4 & 60.4 & 45.2 \\
& & EA        & & 79.1 & 77.6 & 72.6 & 59.4 & 44.2 \\
& & CAOTE     & & 79.0 & 77.9 & 73.2 & 61.2 & 46.7 \\
\rowcolor{oursrow}
& & \textbf{Ours} & & \textbf{79.3} & \textbf{78.0} & \textbf{73.5} & \textbf{68.8} & \textbf{56.7} \\
\midrule

\multirow[t]{18}{*}{\textbf{LLaMA3.1-8B}}
& \multirow{6}{*}{LongBench Avg.} & SnapKV & \multirow{6}{*}{47.3} & 45.8 & 45.2 & 43.4 & 38.2 & 29.5 \\
& & PyramidKV & & 45.8 & 45.3 & 43.7 & 39.2 & 31.6 \\
& & KVzip     & & 46.1 & 45.8 & 44.5 & 40.9 & 35.4 \\
& & EA        & & 46.4 & 46.5 & 46.9 & 43.1 & 36.9 \\
& & CAOTE     & & 46.3 & 46.2 & 46.4 & 43.4 & 37.8 \\
\rowcolor{oursrow}
& & \textbf{Ours} & & \textbf{46.6} & \textbf{47.0} & \textbf{47.0} & \textbf{44.8} & \textbf{41.8} \\
\cmidrule(lr){2-9}

& \multirow{6}{*}{RULER 8K} & SnapKV & \multirow{6}{*}{80.2} & 79.3 & 78.0 & 71.5 & 56.8 & 41.2 \\
& & PyramidKV & & 79.6 & 78.7 & 73.9 & 58.1 & 44.0 \\
& & KVzip     & & 79.8 & 78.6 & 74.3 & 61.2 & 46.8 \\
& & EA        & & 79.7 & 79.1 & 73.5 & 63.7 & 49.6 \\
& & CAOTE     & & 79.9 & 78.9 & 75.2 & 63.1 & 51.0 \\
\rowcolor{oursrow}
& & \textbf{Ours} & & \textbf{80.0} & \textbf{79.5} & \textbf{76.0} & \textbf{73.8} & \textbf{61.2} \\
\cmidrule(lr){2-9}

& \multirow{6}{*}{RULER 16K} & SnapKV & \multirow{6}{*}{75.0} & 74.0 & 72.3 & 64.1 & 45.8 & 28.8 \\
& & PyramidKV & & 74.5 & 72.9 & 66.2 & 47.1 & 31.6 \\
& & KVzip     & & 74.4 & 73.1 & 66.9 & 50.4 & 33.8 \\
& & EA        & & 74.6 & 72.7 & 67.5 & 51.0 & 36.2 \\
& & CAOTE     & & 74.3 & 73.6 & 68.0 & 56.1 & 40.6 \\
\rowcolor{oursrow}
& & \textbf{Ours} & & \textbf{74.8} & \textbf{73.8} & \textbf{70.5} & \textbf{66.3} & \textbf{52.5} \\
\midrule

\multirow[t]{18}{*}{\textbf{Gemma3-12B}}
& \multirow{6}{*}{LongBench Avg.} & SnapKV & \multirow{6}{*}{51.2} & 50.0 & 49.5 & 47.7 & 43.2 & 34.8 \\
& & PyramidKV & & 50.1 & 49.8 & 48.1 & 44.0 & 36.2 \\
& & KVzip     & & 50.3 & 49.9 & 48.4 & 44.8 & 38.0 \\
& & EA        & & 51.0 & 50.4 & 48.1 & 41.2 & 32.5 \\
& & CAOTE     & & 50.5 & 50.0 & 49.0 & 46.3 & 40.2 \\
\rowcolor{oursrow}
& & \textbf{Ours} & & \textbf{51.2} & \textbf{50.8} & \textbf{50.0} & \textbf{48.4} & \textbf{43.5} \\
\cmidrule(lr){2-9}

& \multirow{6}{*}{RULER 8K} & SnapKV & \multirow{6}{*}{88.0} & 87.3 & 86.0 & 79.2 & 63.8 & 45.8 \\
& & PyramidKV & & 87.5 & 86.6 & 81.3 & 65.4 & 47.9 \\
& & KVzip     & & 87.4 & 86.8 & 81.0 & 68.2 & 49.7 \\
& & EA        & & 87.7 & 86.5 & 82.2 & 69.6 & 52.4 \\
& & CAOTE     & & 87.6 & 87.0 & 82.9 & 71.0 & 55.8 \\
\rowcolor{oursrow}
& & \textbf{Ours} & & \textbf{87.8} & \textbf{87.2} & \textbf{83.8} & \textbf{76.0} & \textbf{68.0} \\
\cmidrule(lr){2-9}

& \multirow{6}{*}{RULER 16K} & SnapKV & \multirow{6}{*}{80.0} & 79.2 & 77.8 & 71.4 & 55.6 & 38.7 \\
& & PyramidKV & & 79.5 & 78.4 & 72.9 & 56.5 & 42.6 \\
& & KVzip     & & 79.4 & 78.6 & 74.2 & 63.1 & 49.2 \\
& & EA        & & 79.6 & 78.2 & 73.2 & 62.6 & 47.2 \\
& & CAOTE     & & 79.3 & 78.8 & 74.6 & 62.8 & 49.6 \\
\rowcolor{oursrow}
& & \textbf{Ours} & & \textbf{79.8} & \textbf{79.0} & \textbf{75.5} & \textbf{68.0} & \textbf{60.0} \\
\bottomrule
\end{tabular}
\caption{
Aggregate prefill-time compression results across three model families.
We report LongBench average and RULER results at 8K and 16K context lengths
as a function of the compression ratio \(\rho\), where \(\rho\) denotes the
fraction of evicted KV entries. 
Detailed task-level results for LLaMA3.1-8B and RULER 4K results are provided in
Appendix~\ref{app:experiment_result}. \textbf{Bold} denotes the
best score within each model--metric block, excluding the full-cache reference.
EA denotes Expected Attention.
}
\label{tab:prefill_aggregate_main}
\end{table*}

\definecolor{oursrow}{HTML}{EAF2FB}   
\definecolor{headrow}{HTML}{F2F2F2}   

\begin{table*}[t]
\centering
\scriptsize
\setlength{\tabcolsep}{2.6pt}
\renewcommand{\arraystretch}{1.02}

\resizebox{\textwidth}{!}{%
\begin{tabular}{@{}lllcccccccccccc@{}}
\toprule
\multirow{3}{*}{\textbf{Model}}
& \multirow{3}{*}{\textbf{Data}}
& \multirow{3}{*}{\textbf{Method}}
& \multicolumn{12}{c}{\textbf{Maximum generation length}} \\
\cmidrule(lr){4-15}
& &
& \multicolumn{4}{c}{\textbf{2048}}
& \multicolumn{4}{c}{\textbf{4096}}
& \multicolumn{4}{c}{\textbf{8192}} \\
\cmidrule(lr){4-7} \cmidrule(lr){8-11} \cmidrule(lr){12-15}
\rowcolor{headrow}
& &
& \textbf{No comp.} & \textbf{$2\times$} & \textbf{$4\times$} & \textbf{$12\times$}
& \textbf{No comp.} & \textbf{$2\times$} & \textbf{$4\times$} & \textbf{$12\times$}
& \textbf{No comp.} & \textbf{$2\times$} & \textbf{$4\times$} & \textbf{$12\times$} \\
\midrule

\multirow{12}{*}{\shortstack[c]{DeepSeek-R1\\Qwen-7B}}
& \multirow{6}{*}{MATH500}
& SnapKV
& \multirow{6}{*}{81.6} & 80.6 & 77.2 & 64.8
& \multirow{6}{*}{84.0} & 83.1 & 78.4 & 60.2
& \multirow{6}{*}{84.2} & 83.3 & 78.0 & 58.8 \\
& & PyramidKV
& & 80.4 & 76.8 & 63.9
& & 82.9 & 78.0 & 59.6
& & 83.0 & 77.5 & 57.9 \\
& & EA
&  & 80.9 & 78.7 & 68.0
&  & 83.4 & 80.0 & 65.0
&  & 83.2 & 80.2 & 63.4 \\
& & CAOTE
&  & 81.0 & 79.0 & 67.6
&  & 83.6 & 80.4 & 66.8
&  & 83.7 & 80.6 & 65.2 \\
& & R-KV
&  & \textbf{81.2} & 80.1 & \textbf{76.6}
&  & \textbf{83.8} & 82.2 & 78.6
&  & \textbf{84.0} & 82.8 & 79.2 \\
& & \cellcolor{oursrow}\textbf{Ours}
& \cellcolor{oursrow} & \cellcolor{oursrow}81.1 & \cellcolor{oursrow}\textbf{80.5} & \cellcolor{oursrow}\textbf{76.6}
& \cellcolor{oursrow} & \cellcolor{oursrow}83.7 & \cellcolor{oursrow}\textbf{82.8} & \cellcolor{oursrow}\textbf{79.8}
& \cellcolor{oursrow} & \cellcolor{oursrow}\textbf{84.0} & \cellcolor{oursrow}\textbf{83.0} & \cellcolor{oursrow}\textbf{80.4} \\
\cmidrule(lr){2-15}

& \multirow{6}{*}{AIME25}
& SnapKV
& \multirow{6}{*}{17.8} & 15.6 & 12.2 & 7.8
& \multirow{6}{*}{23.3} & 20.0 & 16.7 & 8.9
& \multirow{6}{*}{27.8} & 23.3 & 17.8 & 10.0 \\
& & PyramidKV
& & 15.6 & 12.2 & 7.8
& & 20.0 & 15.6 & 8.9
& & 23.3 & 17.8 & 10.0 \\
& & EA
&  & 16.7 & 13.3 & 8.9
&  & 21.1 & 17.8 & 11.1
&  & 24.4 & 20.0 & 12.2 \\
& & CAOTE
&  & 15.6 & 14.4 & 10.0
&  & 22.2 & 18.9 & 13.3
&  & 25.6 & 21.1 & 14.4 \\
& & R-KV
&  & \textbf{17.8} & 16.7 & \textbf{14.4}
&  & \textbf{23.3} & 21.1 & \textbf{18.9}
&  & \textbf{27.8} & 25.6 & \textbf{21.1} \\
& & \cellcolor{oursrow}\textbf{Ours}
& \cellcolor{oursrow} & \cellcolor{oursrow}\textbf{17.8} & \cellcolor{oursrow}\textbf{17.8} & \cellcolor{oursrow}13.3
& \cellcolor{oursrow} & \cellcolor{oursrow}22.2 & \cellcolor{oursrow}\textbf{22.2} & \cellcolor{oursrow}17.8
& \cellcolor{oursrow} & \cellcolor{oursrow}26.7 & \cellcolor{oursrow}\textbf{26.7} & \cellcolor{oursrow}20.0 \\
\midrule

\multirow{12}{*}{\shortstack[c]{Qwen2.5\\Math-7B}}
& \multirow{6}{*}{MATH500}
& SnapKV
& \multirow{6}{*}{79.8} & 79.0 & 75.9 & 60.8
& \multirow{6}{*}{82.5} & 81.3 & 76.8 & 56.8
& \multirow{6}{*}{82.9} & 81.4 & 76.3 & 55.2 \\
& & PyramidKV
& & 78.8 & 75.6 & 59.9
& & 81.1 & 76.4 & 56.0
& & 81.2 & 75.8 & 54.6 \\
& & EA
&  & 79.1 & 76.8 & 63.8
&  & 81.7 & 78.2 & 61.8
&  & 81.8 & 78.1 & 60.3 \\
& & CAOTE
&  & 79.3 & 77.6 & 65.4
&  & 82.0 & 79.0 & 63.0
&  & 82.0 & 78.8 & 62.1 \\
& & R-KV
&  & \textbf{79.6} & 78.9 & 72.8
&  & 82.2 & 80.6 & 76.5
&  & \textbf{82.4} & 81.0 & 77.0 \\
& & \cellcolor{oursrow}\textbf{Ours}
& \cellcolor{oursrow} & \cellcolor{oursrow}79.5 & \cellcolor{oursrow}\textbf{79.3} & \cellcolor{oursrow}\textbf{74.0}
& \cellcolor{oursrow} & \cellcolor{oursrow}\textbf{82.3} & \cellcolor{oursrow}\textbf{81.2} & \cellcolor{oursrow}\textbf{78.0}
& \cellcolor{oursrow} & \cellcolor{oursrow}82.3 & \cellcolor{oursrow}\textbf{81.4} & \cellcolor{oursrow}\textbf{79.0} \\
\cmidrule(lr){2-15}

& \multirow{6}{*}{AIME25}
& SnapKV
& \multirow{6}{*}{11.1} & 8.9 & 6.7 & 3.3
& \multirow{6}{*}{15.6} & 12.2 & 8.9 & 4.4
& \multirow{6}{*}{18.9} & 13.3 & 10.0 & 5.6 \\
& & PyramidKV
& & 8.9 & 6.7 & 3.3
& & 12.2 & 8.9 & 4.4
& & 14.4 & 10.0 & 5.6 \\
& & EA
&  & 10.0 & 7.8 & 4.4
&  & 13.3 & 10.0 & 6.7
&  & 15.6 & 12.2 & 7.8 \\
& & CAOTE
&  & 10.0 & 8.9 & 5.6
&  & 14.4 & 11.1 & 7.8
&  & 16.7 & 13.3 & 8.9 \\
& & R-KV
&  & \textbf{11.1} & 10.0 & \textbf{7.8}
&  & \textbf{15.6} & 13.3 & \textbf{11.1}
&  & \textbf{18.9} & 16.7 & \textbf{13.3} \\
& & \cellcolor{oursrow}\textbf{Ours}
& \cellcolor{oursrow} & \cellcolor{oursrow}10.0 & \cellcolor{oursrow}\textbf{11.1} & \cellcolor{oursrow}6.7
& \cellcolor{oursrow} & \cellcolor{oursrow}14.4 & \cellcolor{oursrow}\textbf{14.4} & \cellcolor{oursrow}10.0
& \cellcolor{oursrow} & \cellcolor{oursrow}17.8 & \cellcolor{oursrow}\textbf{17.8} & \cellcolor{oursrow}\textbf{13.3} \\

\bottomrule
\end{tabular}%
}

\caption{
Generation-time compression results under different maximum generation lengths.
Scores are mean pass@1 over three independent stochastic runs.
For \(n\in\{2,4,12\}\), \(n\times\) denotes that the eligible generated-token
KV cache is kept approximately \(n\) times smaller.
\textbf{Bold} denotes the best score within
each setting, excluding the no-compression reference.
}
\label{tab:generation_time}
\end{table*}

\subsection{Experimental Setup}
\paragraph{Models and datasets.}
We evaluate KV cache compression in two inference settings. In
\emph{prefill-time compression}, the prompt KV cache is compressed after
long-context encoding and before answer generation. This setting tests whether
compression preserves task-relevant information stored in the prompt cache. Since
compression is applied after the full prompt forward pass, it reduces the cache
used for subsequent generation but not the compute or transient memory required
for initial prompt prefill. We evaluate Qwen3-8B~\citep{yang2025qwen3technicalreport},
LLaMA3.1-8B~\citep{grattafiori2024llama3herdmodels}, and
Gemma3-12B~\citep{gemmateam2025gemma3technicalreport} on the full LongBench
benchmark~\citep{bai-etal-2024-longbench}. 
We also evaluate on RULER at 4K, 8K, and 16K
context lengths to test controlled long-context retrieval and
aggregation~\citep{hsieh2024rulerwhatsrealcontext}.

In \emph{generation-time compression}, the prompt cache is kept intact and
compression is applied online to generated-token KV entries during decoding.
This setting tests whether cache growth during long reasoning can be controlled
without disrupting generation. We evaluate
DeepSeek-R1-Distill-Qwen-7B~\citep{Guo_2025} and
Qwen2.5-Math-7B~\citep{Yang2024Qwen25MathTR} on
MATH500~\citep{lightman2024lets} and AIME25~\citep{MAA_AIME}, which require
multi-step mathematical reasoning.

\paragraph{Baselines.}
We compare with representative training-free KV cache compression methods:
SnapKV and PyramidKV for attention-based token selection
\citep{DBLP:conf/nips/LiHYVLYCLC24, DBLP:journals/corr/abs-2406-02069},
KVzip for query-agnostic context reconstruction \citep{kim2026kvzip},
Expected Attention for future-attention estimation
\citep{DBLP:journals/corr/abs-2510-00636}, and CAOTE for attention-output
error scoring \citep{goel2026caote}. We also include the no-compression
full-cache model as the reference. For generation-time compression, we
additionally compare with R-KV, a reasoning-specific KV cache compression
baseline \citep{cai2026rkv}. 
KVzip is evaluated only in prefill-time compression, as it targets
query-agnostic prompt-cache reconstruction rather than online pruning.

\paragraph{Implementation and budget matching.}
All methods are evaluated with the same models, datasets, decoding settings,
compression timing, and matched final retained layer--KV-head--token budgets.
We keep each baseline's native pruning granularity, protection rule, and
layer-budget allocation, but protected entries always count toward the retained
budget. Thus, no method receives free protected cache, and $\rho$ denotes the
evicted fraction under this accounting. For our method, each layer-token
decision is shared across KV heads and is counted after expansion to the
corresponding KV-head entries. Appendix~\ref{app:implementation} gives further
details.

\subsection{Prefill-time Compression}
\label{sec:exp_prefilling}

Table~\ref{tab:prefill_aggregate_main} summarizes prefill-time compression
results across models, benchmarks, and context lengths. The table reports
LongBench average performance together with RULER results at 8K and 16K
context lengths under matched cache budgets. As the compression ratio \(\rho\)
increases, proxy-based baselines degrade more rapidly, whereas our method
preserves performance more effectively. This pattern is largely consistent
across the three model families and becomes especially clear under aggressive
compression and longer-context RULER settings.

The results also show the benefit of evaluating evictions in predictive space
rather than relying only on intermediate proxy signals.
CAOTE measures eviction-induced error in attention-output
space. In contrast, our method propagates the estimated residual-stream
perturbation through the final normalization and LM head, forms estimated
compressed-cache logits, and evaluates their KL to the full-cache next-token
distribution. This predictive-space scoring gives modest gains at mild
compression, but becomes more beneficial when the retained cache budget is
tight, especially on RULER 8K and 16K.

\subsection{Generation-time Compression}
\label{sec:generation_time}

Unlike prefill-time compression, generated-token pruning is applied repeatedly
as decoding proceeds. Longer generation horizons therefore create larger caches
and more pruning events, making errors from poor eviction decisions more likely
to accumulate over the reasoning process. Table~\ref{tab:generation_time}
reports pass@1 at different maximum generation lengths under full-cache decoding
and \(2\times\), \(4\times\), and \(12\times\) generated-token cache
compression.

Mild compression causes only small performance changes, but tighter budgets and
longer generation horizons make the choice of eviction rule more consequential.
Attention-based methods and CAOTE degrade more rapidly under these settings,
while R-KV and our method remain more stable. R-KV is a strong
reasoning-specific decoding-time baseline and is best in some settings. It uses
an explicit importance--redundancy trade-off for reasoning traces, whereas our
method uses the same estimated-logit KL score as in prefill-time compression
without reasoning-specific redundancy heuristics. Its competitive
generation-time performance suggests that the proposed behavior-preserving score
transfers beyond one-shot prompt-cache compression to repeated online pruning of
generated-token caches.

At \(12\times\) compression, increasing the maximum generation length from
2048 to 8192 increases SnapKV's MATH500 degradation relative to full-cache
decoding from \(16.8\) to \(25.4\) points on DeepSeek-R1-Distill-Qwen-7B and from
\(19.0\) to \(27.7\) points on Qwen2.5-Math-7B, whereas our degradation remains
within \(3.8\)--\(5.8\) points across these settings.

Appendix~\ref{app:efficiency} quantifies the quality--efficiency trade-off.
At matched retained-KV budgets, our method improves task quality at a higher
compression-time cost than lightweight heuristics. In cross-budget comparisons,
it retains \(2.5\)--\(3\times\) fewer KV entries than SnapKV while achieving
higher task scores and remaining faster than full-cache inference.

\section{Ablation Study}
We ablate the main scoring and selection components in
Appendix~\ref{app:ablation}. 
\paragraph{Final-normalization propagation.}
The ablated variant treats the final normalization as identity, using
\(W\sum_{\ell=1}^{L}
\Delta u_{\ell t}(d_\ell)\) instead of \(WJ_t\sum_{\ell=1}^{L}
\Delta u_{\ell t}(d_\ell)\) to estimate the logit perturbation. The drop across
metrics confirms that final normalization changes how residual-stream
perturbations translate into compressed-cache logits and, consequently, into
output-distribution changes.

\paragraph{Scoring window size.}
Using a single query position gives a noisier estimate of the behavioral effect
of eviction, whereas eight positions provide little improvement over four. We
therefore use \(m_{\mathrm{pre}}=m_{\mathrm{gen}}=4\) by default.

\paragraph{Top-\(K\) KL evaluation.}
We evaluate the estimated-logit KL score on the renormalized top-\(K\) support
of the full-cache distribution. The small gains from increasing \(K\) suggest
that a compact output support captures most of the predictive signal while
keeping scoring efficient.

\paragraph{Selection refinement.}
Pool-refine uses singleton scores only to form a candidate pool, and selects
final evictions by reevaluating the marginal KL cost under the current
multi-entry mask. Unlike fixed additive scores, this refinement accounts for
mask-level interactions caused by attention renormalization and changes in the
estimated compressed-cache distribution. Its gains over singleton selection
suggest that these interactions matter under tight cache budgets.

\section{Conclusion}
We presented Behavior-Preserving KV Cache Compression, a training-free
framework that recasts KV eviction as preserving the predictive behavior of the
full-cache model rather than optimizing proxy importance signals. Our method
scores candidate evictions by estimating the compressed-cache logits induced by
their removal and evaluating their KL to the full-cache next-token distribution,
using only pre-eviction forward statistics and no candidate-specific masked
forward passes. Across both prefill-time and generation-time compression,
behavior-preserving scoring maintains task performance under tight cache
budgets, supporting output-distribution preservation as an effective principle
for training-free KV cache compression.

\section*{Limitations}

A common challenge in training-free KV cache compression is that eviction decisions must be made before the future decoding queries that will read the cache are known. Our method follows this setting and uses the next-token distributions at available query positions as the observable reference behavior. 
This design preserves behavior only with respect to the available compression-time queries, and does not provide a guarantee for arbitrary unseen future continuations, where later queries may amplify or dampen the effect of eviction. 
Developing compression-time signals that better anticipate unseen future queries remains an important direction for KV cache compression.
\section*{Acknowledgments}
This work was the result of project supported by KT(Korea Telecom).
This work was also supported by Institute for Information \& communications
Technology Planning \& Evaluation (IITP) grants funded by the Korea government
(MSIT): No. RS-2024-00457882 (AI Research Hub Project),
RS-2019-II190075 (Artificial Intelligence Graduate School Program (KAIST)),
No. RS-2024-00397310 (Development of an AI Simulator for Creating Transparent
Compounds that Can Be Altered for Tactile Sensation), and
No. RS-2024-00343989 (Enhancing the Ethics of Data Characteristics and
Generation AI Models for Social and Ethical Learning).
This work was supported by the IITP (Institute of Information \&
Communications Technology Planning \& Evaluation)-ITRC (Information Technology
Research Center) grant funded by the Korea government (Ministry of Science
and ICT) (IITP-2026-RS-2024-00436857).
This research was supported by the ``Advanced GPU Utilization Support
Program'' funded by the Government of the Republic of Korea
(Ministry of Science and ICT).

\bibliography{custom}

\appendix
\clearpage
\onecolumn
\section{Derivation of the Layer Perturbations from KV Eviction}
\label{app:attention_perturbation_derivation}

Fix a layer \(\ell\), head \(r\in\{1,\ldots,H\}\), and scoring query position
\(t\in\mathcal T\). Let \(\mathcal I_t\subseteq\{1,\ldots,S\}\) denote the cache
positions visible to query \(t\) under the causal mask. Equivalently, one may
sum over all positions by setting \(a_{\ell rti}=0\) for positions not visible
to \(t\).

Recall that
\[
\tilde v_{\ell rj}=W^O_{\ell r}v_{\ell rj}
\]
is the output-projected value vector for token \(j\) at layer \(\ell\) and head
\(r\). The full-cache projected head output is
\[
o_{\ell rt}
=
\sum_{i\in\mathcal I_t}
a_{\ell rti}\tilde v_{\ell ri}.
\]

For a layer retention mask \(d_\ell\), let \(\bar d_{\ell j}=1-d_{\ell j}\).
The removed attention mass and removed projected contribution are
\[
A_{\ell rt}(d_\ell)
=
\sum_{j\in\mathcal I_t}
\bar d_{\ell j}a_{\ell rtj},
\qquad
R_{\ell rt}(d_\ell)
=
\sum_{j\in\mathcal I_t}
\bar d_{\ell j}a_{\ell rtj}\tilde v_{\ell rj}.
\]

In the deletion counterfactual, the pre-eviction query, keys, and values at
layer \(\ell\) are held fixed. Removing masked entries deletes their attention
terms and renormalizes the retained attention weights by the retained mass
\(1-A_{\ell rt}(d_\ell)\). Assuming \(1-A_{\ell rt}(d_\ell)>0\), the masked
projected head output is
\[
o_{\ell rt}^{(d_\ell)}
=
\sum_{i\in\mathcal I_t}
\frac{d_{\ell i}a_{\ell rti}}{1-A_{\ell rt}(d_\ell)}
\tilde v_{\ell ri}
=
\frac{
o_{\ell rt}-R_{\ell rt}(d_\ell)
}{
1-A_{\ell rt}(d_\ell)
}.
\]
Therefore the head-level perturbation is
\[
\Delta o_{\ell rt}(d_\ell)
=
o_{\ell rt}^{(d_\ell)}-o_{\ell rt}
=
\frac{
A_{\ell rt}(d_\ell)o_{\ell rt}
-
R_{\ell rt}(d_\ell)
}{
1-A_{\ell rt}(d_\ell)
}.
\]
Substituting the definitions of \(A_{\ell rt}(d_\ell)\) and
\(R_{\ell rt}(d_\ell)\) gives the equivalent entrywise form
\[
\Delta o_{\ell rt}(d_\ell)
=
\frac{
\sum_{j\in\mathcal I_t}
\bar d_{\ell j}a_{\ell rtj}
\left(
o_{\ell rt}-\tilde v_{\ell rj}
\right)
}{
1-A_{\ell rt}(d_\ell)
}.
\]

Summing over attention heads gives the residual-stream perturbation injected by
the attention block at layer \(\ell\):
\[
\Delta u_{\ell t}(d_\ell)
=
\sum_{r=1}^{H}
\Delta o_{\ell rt}(d_\ell).
\]
This expression is exact for the deletion counterfactual in which the
pre-eviction query, keys, and values at layer \(\ell\) are held fixed. Responses
of later attention, MLP, and normalization layers are treated as approximation
error in the main residual-path surrogate.

\clearpage
\twocolumn
\section{Validation of the Residual-Path Approximation}
\label{app:residual_path_validation}

We validate whether the residual-path approximation provides a useful ranking
signal under multi-entry KV eviction. This setting matches the actual
compression procedure, where a mask removes many KV entries simultaneously. We
use LLaMA3.1-8B on LongBench examples.

The residual-path approximation is motivated by the additive structure of
Transformer residual connections, where attention blocks write updates into a
shared residual stream
\citep{DBLP:conf/nips/VaswaniSPUJGKP17, elhage2021mathematical}.
We keep this directly written component as a tractable proxy for the
candidate-induced final-state perturbation. This does not assume zero response
from subsequent layers; those responses are treated as approximation error and
evaluated empirically below.

\begin{figure}[t]
\centering
\includegraphics[width=0.48\textwidth]{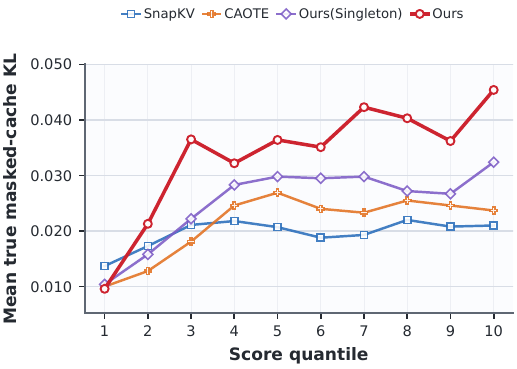}
\caption{
Validation of the residual-path approximation under multi-entry KV eviction on
LLaMA3.1-8B using LongBench. We rank the same eviction masks by evicted SnapKV
attention score, CAOTE attention-output error, our additive singleton KL score,
and our mask-level estimated-logit KL score. For each score quantile, we report
the mean true masked-cache KL \(\mathcal J_{\mathrm{true}}(d)\) from a
masked-cache forward pass. The mask-level estimated-logit KL score gives the
strongest ranking signal, with Spearman correlations of \(0.45\), \(0.53\),
\(0.67\), and \(0.77\) for SnapKV, CAOTE, singleton, and mask-level
estimated-logit KL, respectively.
}
\label{fig:residual_path_validation}
\end{figure}

For an eviction mask \(d\in\mathcal D_B\), we compute the exact masked-cache
behavioral change by running a masked-cache forward pass and measuring
\begin{equation}
\mathcal J_{\mathrm{true}}(d)
=
\frac{1}{|\mathcal T|}
\sum_{t\in\mathcal T}
D_{\mathrm{KL}}\!\left(
p_t
\,\middle\|\,
q_t^{(d)}
\right),
\label{eq:mask_true_kl_validation}
\end{equation}
where \(q_t^{(d)}\) is the predictive distribution after applying the mask. This
quantity includes simultaneous KV removals, attention renormalization, and
propagation through later layers, and serves as the masked-cache reference.

Our residual-path approximation estimates the logit perturbation as
\begin{equation}
\widehat{\Delta z}_{t,d}
=
WJ_t
\sum_{\ell=1}^{L}
\Delta u_{\ell t}(d_\ell),
\label{eq:mask_logit_validation}
\end{equation}
where \(J_t=\nabla N_f(x_t^L)\) is the final-normalization Jacobian at the
full-cache residual state, and \(\Delta u_{\ell t}(d_\ell)\) is the
residual-stream perturbation injected by layer \(\ell\). We then form the
estimated compressed-cache logits
\[
\hat z_{t,d}=z_t+\widehat{\Delta z}_{t,d}
\]
and evaluate the same behavior-preserving KL objective on these estimated
logits:
\begin{equation}
\begin{aligned}
\widehat{\mathcal J}(d)
&=
\frac{1}{|\mathcal T|}
\sum_{t\in\mathcal T}
D_{\mathrm{KL}}\!\left(
p_t
\,\middle\|\,
\hat q_t^{(d)}
\right), \\
\hat q_t^{(d)}
&=
\mathrm{softmax}(\hat z_{t,d}).
\end{aligned}
\label{eq:mask_surrogate_validation}
\end{equation}
Unlike \(\mathcal J_{\mathrm{true}}(d)\), this score uses only full-cache
statistics and requires no masked-cache forward pass. It differs from the true
masked-cache KL only in replacing the exact compressed-cache logits with
residual-path estimated logits.

To contextualize this approximation, we compare with three simpler signals. All
validation scores are evaluated on the same layer-token eviction masks. When a
baseline provides head-specific scores, we aggregate those scores over heads
under the shared layer-token mask. Recall that \(\bar d_{\ell j}=1-d_{\ell j}\)
indicates whether entry \((\ell,j)\) is evicted.

The first signal is the evicted SnapKV attention score,
\begin{equation}
E_{\mathrm{SnapKV}}(d)
=
\frac{1}{LH}
\sum_{\ell=1}^{L}
\sum_{r=1}^{H}
\sum_{j=1}^{S}
\bar d_{\ell j}
s^{\mathrm{SnapKV}}_{\ell rj}.
\label{eq:snapkv_validation_score}
\end{equation}
The second is the CAOTE attention-output error induced by the same mask,
\begin{equation}
E_{\mathrm{CAOTE}}(d)
=
\frac{1}{|\mathcal T|LH}
\sum_{t\in\mathcal T}
\sum_{\ell=1}^{L}
\sum_{r=1}^{H}
\left\|
\Delta C_{\ell rt}(d_\ell)
\right\|_2 .
\label{eq:caote_validation_score}
\end{equation}
The third is an additive singleton variant of our estimated-logit KL score:
\begin{equation}
\begin{aligned}
E_{\mathrm{single}}(d)
&=
\frac{1}{B|\mathcal T|}
\sum_{\ell=1}^{L}
\sum_{j=1}^{S}
\sum_{t\in\mathcal T}
\bar d_{\ell j} \\
&\quad \times
D_{\mathrm{KL}}\!\left(
p_t
\,\middle\|\,
\mathrm{softmax}(\hat z_{t,(\ell,j)})
\right).
\end{aligned}
\label{eq:singleton_validation_score}
\end{equation}
where \(\hat z_{t,(\ell,j)}\) denotes the estimated logits under the singleton
eviction of entry \((\ell,j)\). These comparisons isolate the effect of moving
from attention scores to attention-output errors, from local errors to
predictive-distribution scoring, and from additive singleton evaluation to
mask-level evaluation.

For each compression ratio \(\rho\), we evaluate all scores on the same
multi-entry masks. We rank masks by each score, divide them into quantiles, and
report the mean exact masked-cache KL \(\mathcal J_{\mathrm{true}}(d)\) in each
quantile, together with Spearman rank correlation.
Figure~\ref{fig:residual_path_validation} shows that the mask-level
estimated-logit KL score has the strongest monotonic relationship with exact
masked-cache KL. This supports the residual-path approximation as a proxy for
ranking multi-entry cache-retention masks.
\section{Implementation Details}
\label{app:implementation}

\paragraph{Statistics collected from the full-cache pass.}
Our method uses statistics derived from the unmasked pre-eviction computation
and does not run candidate-specific masked forward passes. For each scoring
position \(t\in\mathcal T\), we obtain the full-cache logits \(z_t\), keep the
top-\(K\) support \(\mathcal K_t\) and the renormalized top-\(K\)
probabilities used by the KL evaluation, and capture the final
pre-normalization residual state \(x_t^L\). For each layer, we recompute only
the attention rows for \(\mathcal T\) from cached keys and query states,
storing \(H\times |\mathcal T|\times S\) attention rows per layer rather than
full \(H\times S\times S\) attention maps. We also store the projected
full-cache head outputs \(o_{\ell rt}\).

The projected value
\[
\tilde v_{\ell rj}=W^O_{\ell r}v_{\ell rj}
\]
denotes the output-projected value contribution used in the derivation. We do
not precompute or store \(\tilde v_{\ell rj}\) for all layers, heads, and
tokens. In implementation, these products are computed for candidate token
chunks during scoring and refinement, and the block-level projected-value
buffers are discarded after use.

\paragraph{Final-normalization Jacobian-vector product.}
Since the LM head is applied after final normalization, the residual-stream
perturbation must be propagated through the final normalization. For
RMSNorm-based models, this Jacobian-vector product can be computed in closed
form. Let
\[
N_f(x)=\gamma\odot \frac{x}{s(x)},
\qquad
s(x)=\sqrt{\frac{\|x\|_2^2}{D}+\epsilon_{\mathrm{rms}}},
\]
where \(D\) is the hidden dimension. For a perturbation \(v\in\mathbb R^D\) at
scoring position \(t\), with \(s_t=s(x_t^L)\), we compute
\[
J_t v
=
\gamma\odot
\left(
\frac{v}{s_t}
-
\frac{x_t^L((x_t^L)^\top v)}{D s_t^3}
\right).
\]
For models without final normalization, we set \(N_f=I\) and \(J_t=I\). The
LM-head bias, if present, does not appear in the perturbation because it
cancels when taking logit differences.

\paragraph{Top-\(K\) KL evaluation.}
The practical score in Eq.~\eqref{eq:estimated_behavior_objective} evaluates
the KL between the full-cache prediction and the estimated compressed-cache
prediction. Full-vocabulary evaluation is expensive, so we evaluate this score
on the top-\(K\) support of the full-cache distribution. Let \(\mathcal K_t\)
be the top-\(K\) tokens under \(p_t=\mathrm{softmax}(z_t)\), and define
\[
\widetilde p_{t,i}
=
\frac{p_{t,i}}{\sum_{k\in\mathcal K_t}p_{t,k}},
\qquad i\in\mathcal K_t .
\]
For a mask \(d\), let
\[
v_{t,d}
=
\sum_{\ell=1}^{L}
\Delta u_{\ell t}(d_\ell)
\]
be the residual-path perturbation before final normalization. Its induced
top-\(K\) logit shift is
\[
\delta_{t,i}(d)
=
W_iJ_tv_{t,d},
\qquad i\in\mathcal K_t ,
\]
where \(W_i\) is the \(i\)-th row of the LM head. Since the estimated
compressed-cache logits are
\(\hat z_{t,d}=z_t+\widehat{\Delta z}_{t,d}\), the estimated compressed-cache
distribution restricted to \(\mathcal K_t\) can be written as
\[
\widetilde q_{t,i}^{(d)}
=
\frac{
\widetilde p_{t,i}\exp(\delta_{t,i}(d))
}{
\sum_{k\in\mathcal K_t}
\widetilde p_{t,k}\exp(\delta_{t,k}(d))
},
\qquad i\in\mathcal K_t .
\]
The implemented top-\(K\) behavior-preserving score is
\begin{equation}
\widehat{\mathcal J}_K(d)
=
\frac{1}{|\mathcal T|}
\sum_{t\in\mathcal T}
D_{\mathrm{KL}}
\!\left(
\widetilde p_t
\,\middle\|\,
\widetilde q_t^{(d)}
\right).
\label{eq:topk_kl_score}
\end{equation}
Equivalently, this KL can be evaluated without explicitly forming
\(\widetilde q_t^{(d)}\):
\begin{align*}
D_{\mathrm{KL}}
\!\left(
\widetilde p_t
\,\middle\|\,
\widetilde q_t^{(d)}
\right)
&=
\log
\sum_{i\in\mathcal K_t}
\widetilde p_{t,i}\exp(\delta_{t,i}(d))\\
&\quad-
\sum_{i\in\mathcal K_t}
\widetilde p_{t,i}\delta_{t,i}(d).
\end{align*}
We evaluate this expression using a numerically stable log-sum-exp. When
\(\mathcal K_t\) is the full vocabulary, Eq.~\eqref{eq:topk_kl_score} recovers
the estimated-logit KL in Eq.~\eqref{eq:estimated_behavior_objective}.

\begin{table*}[t]
\centering
\setlength{\tabcolsep}{4pt}
\renewcommand{\arraystretch}{0.95}
\begin{tabular}{lcccccc}
\toprule
Setting & \(m\) & \(K\) & \(\alpha\) & \(b_{\mathrm{ref}}\) & \(\epsilon\) & Extra \\
\midrule
Prefill-time
& \(m_{\mathrm{pre}}=4\)
& \(K_{\mathrm{pre}}=128\)
& \(\alpha_{\mathrm{pre}}=2\)
& \(b_{\mathrm{pre}}=512\)
& \(10^{-4}\)
& -- \\
Generation-time
& \(m_{\mathrm{gen}}=4\)
& \(K_{\mathrm{gen}}=256\)
& \(\alpha_{\mathrm{gen}}=2\)
& \(b_{\mathrm{gen}}=256\)
& \(10^{-4}\)
& \(I_{\mathrm{comp}}=512,\; w_{\mathrm{recent}}=4\) \\
\bottomrule
\end{tabular}
\caption{
Default implementation hyperparameters for prefill-time and generation-time
compression. \(K\) is the top-\(K\) support size used for KL evaluation. Values
are fixed within each regime and are not tuned per benchmark.
}
\label{tab:default_hyperparams}
\end{table*}

\paragraph{Efficient evaluation during refinement.}
During the refinement stage, we evaluate
\(\widehat{\mathcal J}_K(d_{\mathcal M})\) and
\(\widehat{\mathcal J}_K(d_{\mathcal M\cup\{a\}})\) using the full-cache
attention weights and cached values, rather than running masked forward passes.
Here, \(d_{\mathcal M}\) denotes the mask that evicts the entries in the current
selected set \(\mathcal M\). For a tentative candidate \(a=(\ell,j)\), only the
layer-\(\ell\) removed-mass and removed-contribution statistics change; the
statistics for all other layers are reused. We then compute
\[
\mathrm{cost}(a;\mathcal M)
=
\widehat{\mathcal J}_K(d_{\mathcal M\cup\{a\}})
-
\widehat{\mathcal J}_K(d_{\mathcal M}).
\]
After selecting candidates in a refinement round, we update the current
retention mask and its cached layerwise statistics.

\paragraph{Pool-and-refine mask construction.}
We form \(\mathcal C_{\mathrm{pool}}\) from the
\(\lceil \alpha B\rceil\) eligible candidates with the lowest singleton-eviction
scores. Starting from \(\mathcal M=\emptyset\), each refinement round evaluates
\(\mathrm{cost}(a;\mathcal M)\) for the remaining candidates in
\(\mathcal C_{\mathrm{pool}}\), adds the lowest-cost
\(\min(b_{\mathrm{ref}}, B-|\mathcal M|)\) candidates, and updates the cached
mask statistics. The process stops when \(|\mathcal M|=B\). The cases
\(b_{\mathrm{ref}}=1\) and \(b_{\mathrm{ref}}\ge B\) recover fully greedy
refinement and one-shot selection, respectively.

\paragraph{Numerical stability.}
During scoring and refinement, we reject any candidate whose addition would
make the retained attention mass too small for any scoring position, layer, or
head. Let
\[
A_{\ell rt}(d_\ell)
=
\sum_{i=1}^{S}
(1-d_{\ell i})a_{\ell rti}
\]
be the removed attention mass. Since the pre-eviction attention weights sum to
one over visible positions, the retained attention mass is
\(1-A_{\ell rt}(d_\ell)\). We reject a candidate if
\[
1-A_{\ell rt}(d_\ell) < \epsilon .
\]
This avoids unstable attention renormalization in
Eq.~\eqref{eq:head_perturbation}.

\paragraph{Grouped-query attention.}
For grouped-query or multi-query attention, \(v_{\ell rj}\) denotes the value
vector from the KV head associated with query head \(r\). We compute
Eq.~\eqref{eq:layer_perturbation} at the query-head level using the
corresponding shared KV head, while the cache-retention mask remains shared
across heads within each layer.

\paragraph{Default hyperparameters.}
We use the same behavior-preserving score in both compression regimes, but use
separate implementation hyperparameters because the eligible candidates and
compression schedules differ. In prefill-time compression, compression is
applied once after prompt prefill, \(\mathcal T\) contains the last
\(m_{\mathrm{pre}}\) prompt positions, and the budget is assigned over eligible
prompt layer-token entries. In generation-time compression, prompt entries are
always protected, compression is triggered every \(I_{\mathrm{comp}}\)
generated tokens, and only generated-token entries outside the recent protected
window \(w_{\mathrm{recent}}\) are eligible; \(\mathcal T\) contains the most
recent \(m_{\mathrm{gen}}\) generated positions at each compression point.

Table~\ref{tab:default_hyperparams} lists the top-\(K\) size used for KL
evaluation, pool factor \(\alpha\), refinement batch size \(b_{\mathrm{ref}}\),
and stability threshold \(\epsilon\). For \(n\times\) generation-time
compression, the eligible generated-token cache is kept approximately \(n\)
times smaller than the uncompressed eligible cache. All hyperparameters are
fixed within each regime and are not tuned per benchmark.

\paragraph{Baseline settings and budget matching.}
Unless otherwise stated, baseline hyperparameters follow the settings
recommended in the original papers and their corresponding KVPress
implementations.\footnote{\url{https://github.com/NVIDIA/kvpress}}
We evaluate each method with its native scoring rule, pruning granularity, and
protection rule, and match methods by the final retained
layer--KV-head--token budget.

For a cache of length \(S\), we compute the prefill-time compression ratio by
counting retained layer--KV-head--token entries:
\[
\rho =
1 -
\frac{
\sum_{\ell=1}^{L}\sum_{h=1}^{H_{\mathrm{kv}}}
|\mathcal{S}_{\ell,h}^{\mathrm{keep}}|
}{
L H_{\mathrm{kv}} S
}.
\]
Here, \(\mathcal{S}_{\ell,h}^{\mathrm{keep}}\) denotes the retained token
positions at layer \(\ell\) and KV head \(h\). For head-wise methods, this set
may differ across KV heads. For our method, the layer-token mask is shared
across KV heads within each layer, so a retained or evicted token position
applies to all corresponding KV-head entries in that layer. Protected entries
are counted as retained entries and therefore consume cache budget; no method
receives uncounted protected cache.

For our method, we use \(n_{\mathrm{sink}}=4\) in all prefill-time experiments.
SnapKV and PyramidKV use their observation windows for attention-based scoring
and retain the observation-window entries according to their KVPress
implementations.

\onecolumn

\begin{table*}[t]
\centering
\small
\setlength{\tabcolsep}{4pt}
\begin{tabular}{l l l l}
\toprule
Method & Native granularity & Native protected set & Layer budget \\
\midrule
Ours & Layer-token & Sink & Global \\
SnapKV & KV-head-token & Observation window & Uniform \\
PyramidKV & KV-head-token & Observation window & Pyramidal \\
Expected Attn. & KV-head-token & Sink & Uniform \\
KVzip & KV-head-token & Sink & Global \\
CAOTE & KV-head-token & Sink & Global \\
\bottomrule
\end{tabular}
\caption{
Implementation details for prefill-time compression. Baselines are evaluated
with their original pruning granularity, protection rule, and layer-budget
allocation. Protected entries are included in the matched final retained
layer--KV-head--token budget for all methods, so no method receives uncounted
protected cache.
}
\label{tab:baseline_impl_prefill}
\end{table*}

\section{Experiment Results}
\label{app:experiment_result}
\paragraph{Task-level prefill-time compression results}
\definecolor{oursrow}{HTML}{EAF2FB}   
\definecolor{headrow}{HTML}{F2F2F2}   
\definecolor{catrow}{HTML}{F7F7F7}    

\begingroup
\small
\setlength{\tabcolsep}{3.8pt}
\renewcommand{\arraystretch}{1.03}

\begin{xltabular}{\linewidth}{@{}
>{\raggedright\arraybackslash}p{3.35cm}
>{\raggedright\arraybackslash}p{1.65cm}
c c c c c c
@{}}

\toprule
\rowcolor{headrow}
\textbf{Task} & \textbf{Method} & \textbf{Full}
& $\rho\!=\!0.10$ & $\rho\!=\!0.25$ & $\rho\!=\!0.50$
& $\rho\!=\!0.75$ & $\rho\!=\!0.90$ \\
\midrule
\endfirsthead

\toprule
\rowcolor{headrow}
\textbf{Task} & \textbf{Method} & \textbf{Full}
& $\rho\!=\!0.10$ & $\rho\!=\!0.25$ & $\rho\!=\!0.50$
& $\rho\!=\!0.75$ & $\rho\!=\!0.90$ \\
\midrule
\endhead

\midrule
\multicolumn{8}{r}{\textit{Continued on next page}} \\
\endfoot

\bottomrule
\endlastfoot

\rowcolor{catrow}
\multicolumn{8}{l}{\textbf{Single-document QA}} \\
\midrule

\textbf{NarrativeQA} & SnapKV       & \multirow[c]{6}{*}{28.8} & 27.2 & 27.3 & 25.5 & 22.9 & 17.5 \\
 & PyramidKV    &  & 27.3 & 27.4 & 25.9 & 23.4 & 19.3 \\
 & KVzip        &  & \underline{27.7} & 27.5 & 26.6 & 24.8 & 21.0 \\
 & EA           &  & \underline{27.7} & \textbf{28.3} & \textbf{28.4} & 26.2 & 22.6 \\
 & CAOTE        &  & \underline{27.7} & 27.6 & 28.1 & \underline{26.9} & \underline{23.0} \\
\rowcolor{oursrow}
 & \textbf{Ours} &  & \textbf{28.2} & \underline{28.1} & \underline{28.2} & \textbf{27.4} & \textbf{25.4} \\
\cmidrule(l){1-8}

\textbf{Qasper} & SnapKV       & \multirow[c]{6}{*}{36.0} & 34.6 & 34.3 & 33.3 & 30.1 & 24.7 \\
 & PyramidKV    &  & 34.4 & 34.6 & 33.3 & 30.3 & 26.8 \\
 & KVzip        &  & 35.0 & 34.9 & 34.3 & 31.6 & 28.4 \\
 & EA           &  & \textbf{35.6} & 35.0 & \textbf{36.0} & 33.2 & 29.6 \\
 & CAOTE        &  & 35.1 & \underline{35.4} & 35.7 & \underline{33.8} & \underline{30.3} \\
\rowcolor{oursrow}
 & \textbf{Ours} &  & \underline{35.4} & \textbf{35.6} & \underline{35.8} & \textbf{34.2} & \textbf{32.8} \\
\cmidrule(l){1-8}

\textbf{MultiFieldQA-en} & SnapKV       & \multirow[c]{6}{*}{55.0} & 53.6 & 53.3 & 51.8 & 48.8 & 43.7 \\
 & PyramidKV    &  & 53.6 & 53.1 & 52.2 & 49.4 & 45.7 \\
 & KVzip        &  & 54.2 & 53.5 & 52.9 & 50.6 & 47.6 \\
 & EA           &  & \underline{54.4} & \underline{54.1} & \textbf{55.0} & 52.4 & 48.2 \\
 & CAOTE        &  & 54.2 & 54.0 & 54.3 & \underline{52.8} & \underline{49.3} \\
\rowcolor{oursrow}
 & \textbf{Ours} &  & \textbf{54.5} & \textbf{54.6} & \underline{54.8} & \textbf{53.6} & \textbf{51.4} \\
\cmidrule(l){1-8}

\textbf{MultiFieldQA-zh} & SnapKV       & \multirow[c]{6}{*}{42.5} & 40.8 & 40.4 & 39.2 & 36.5 & 31.4 \\
 & PyramidKV    &  & 41.4 & 40.8 & 39.9 & 37.3 & 32.7 \\
 & KVzip        &  & 41.1 & 40.9 & 40.4 & 38.4 & 35.2 \\
 & EA           &  & 41.5 & \underline{42.1} & \textbf{42.7} & 40.2 & 35.8 \\
 & CAOTE        &  & \underline{41.8} & 41.3 & 42.3 & \underline{40.5} & \underline{37.3} \\
\rowcolor{oursrow}
 & \textbf{Ours} &  & \textbf{41.9} & \textbf{42.3} & \underline{42.5} & \textbf{41.3} & \textbf{38.9} \\
\midrule

\rowcolor{catrow}
\multicolumn{8}{l}{\textbf{Multi-document QA}} \\
\midrule

\textbf{HotpotQA} & SnapKV       & \multirow[c]{6}{*}{58.8} & 57.2 & 56.5 & 54.6 & 48.6 & 40.0 \\
 & PyramidKV    &  & 57.4 & 56.7 & 54.5 & 49.8 & 42.5 \\
 & KVzip        &  & 57.8 & 57.2 & 55.7 & 51.7 & 46.1 \\
 & EA           &  & 57.7 & \underline{58.1} & \underline{58.7} & 55.5 & 48.9 \\
 & CAOTE        &  & \underline{57.9} & 57.6 & 58.5 & \underline{55.8} & \underline{50.0} \\
\rowcolor{oursrow}
 & \textbf{Ours} &  & \textbf{58.4} & \textbf{58.7} & \textbf{58.8} & \textbf{56.8} & \textbf{53.8} \\
\cmidrule(l){1-8}

\textbf{2WikiMQA} & SnapKV       & \multirow[c]{6}{*}{54.5} & 52.6 & 52.2 & 50.5 & 44.6 & 35.7 \\
 & PyramidKV    &  & 53.1 & 52.7 & 50.7 & 45.8 & 38.3 \\
 & KVzip        &  & 53.1 & 52.8 & 51.5 & 47.8 & 41.9 \\
 & EA           &  & \underline{53.9} & \underline{54.0} & \textbf{54.7} & 51.1 & 44.8 \\
 & CAOTE        &  & 53.7 & 53.5 & \underline{54.4} & \underline{51.3} & \underline{45.9} \\
\rowcolor{oursrow}
 & \textbf{Ours} &  & \textbf{54.1} & \textbf{54.4} & \underline{54.4} & \textbf{52.2} & \textbf{49.8} \\
\cmidrule(l){1-8}

\textbf{MuSiQue} & SnapKV       & \multirow[c]{6}{*}{38.0} & 36.5 & 35.8 & 34.1 & 28.2 & 19.7 \\
 & PyramidKV    &  & 36.6 & 35.8 & 33.9 & 29.0 & 21.5 \\
 & KVzip        &  & 37.0 & 36.2 & 34.7 & 31.2 & 25.3 \\
 & EA           &  & 37.0 & 37.0 & \textbf{38.4} & 34.3 & 27.9 \\
 & CAOTE        &  & \underline{37.2} & \underline{37.1} & 37.6 & \underline{35.3} & \underline{29.4} \\
\rowcolor{oursrow}
 & \textbf{Ours} &  & \textbf{37.3} & \textbf{38.0} & \underline{38.0} & \textbf{36.4} & \textbf{33.0} \\
\cmidrule(l){1-8}

\textbf{DuReader} & SnapKV       & \multirow[c]{6}{*}{36.5} & 35.0 & 34.3 & 32.5 & 26.4 & 18.2 \\
 & PyramidKV    &  & 35.2 & 34.3 & 32.6 & 27.7 & 20.3 \\
 & KVzip        &  & \underline{35.6} & 35.1 & 33.2 & 29.6 & 24.1 \\
 & EA           &  & 35.3 & \underline{35.8} & \textbf{36.9} & 33.3 & 26.8 \\
 & CAOTE        &  & 35.2 & 35.7 & \underline{36.4} & \underline{33.6} & \underline{27.9} \\
\rowcolor{oursrow}
 & \textbf{Ours} &  & \textbf{35.7} & \textbf{36.5} & 36.3 & \textbf{34.5} & \textbf{31.6} \\
\midrule

\rowcolor{catrow}
\multicolumn{8}{l}{\textbf{Summarization}} \\
\midrule

\textbf{GovReport} & SnapKV       & \multirow[c]{6}{*}{33.2} & 32.1 & 31.8 & 30.9 & 28.3 & 24.1 \\
 & PyramidKV    &  & 31.9 & 31.6 & 30.8 & 28.7 & 24.9 \\
 & KVzip        &  & \underline{32.3} & 31.9 & 31.2 & 29.6 & 27.3 \\
 & EA           &  & \underline{32.3} & \underline{32.7} & \textbf{33.2} & 30.9 & 28.1 \\
 & CAOTE        &  & 32.0 & 32.5 & 32.4 & \underline{31.0} & \underline{28.7} \\
\rowcolor{oursrow}
 & \textbf{Ours} &  & \textbf{32.6} & \textbf{32.9} & \underline{33.0} & \textbf{31.6} & \textbf{30.7} \\
\cmidrule(l){1-8}

\textbf{QMSum} & SnapKV       & \multirow[c]{6}{*}{26.0} & 24.9 & 24.4 & 23.5 & 21.3 & 17.3 \\
 & PyramidKV    &  & \underline{25.1} & 24.2 & 23.6 & 21.4 & 17.8 \\
 & KVzip        &  & \underline{25.1} & 24.5 & 24.5 & 22.7 & 20.3 \\
 & EA           &  & 25.0 & \underline{25.6} & \textbf{25.8} & \underline{24.2} & 21.3 \\
 & CAOTE        &  & \underline{25.1} & 25.4 & \underline{25.5} & 24.0 & \underline{21.7} \\
\rowcolor{oursrow}
 & \textbf{Ours} &  & \textbf{25.2} & \textbf{25.8} & \textbf{25.8} & \textbf{24.6} & \textbf{23.3} \\
\cmidrule(l){1-8}

\textbf{MultiNews} & SnapKV       & \multirow[c]{6}{*}{28.4} & 27.3 & 26.8 & 26.2 & 23.4 & 19.7 \\
 & PyramidKV    &  & 27.1 & 26.7 & 26.1 & 24.1 & 20.6 \\
 & KVzip        &  & 27.4 & 26.9 & 26.5 & 24.7 & 22.5 \\
 & EA           &  & \underline{27.8} & \underline{27.6} & \textbf{28.2} & 26.4 & 23.2 \\
 & CAOTE        &  & 27.5 & 27.2 & 27.7 & \underline{26.6} & \underline{23.6} \\
\rowcolor{oursrow}
 & \textbf{Ours} &  & \textbf{27.9} & \textbf{27.8} & \underline{28.0} & \textbf{27.0} & \textbf{25.3} \\
\cmidrule(l){1-8}

\textbf{VCSUM} & SnapKV       & \multirow[c]{6}{*}{17.2} & 16.1 & 15.5 & 14.7 & 12.2 & 8.4 \\
 & PyramidKV    &  & 16.0 & 15.8 & 14.9 & 12.7 & 9.1 \\
 & KVzip        &  & 16.0 & 16.0 & 15.4 & 13.8 & 11.5 \\
 & EA           &  & \underline{16.3} & \underline{16.5} & \textbf{16.8} & 15.0 & 12.3 \\
 & CAOTE        &  & \underline{16.3} & 16.2 & 16.6 & \underline{15.2} & \underline{12.7} \\
\rowcolor{oursrow}
 & \textbf{Ours} &  & \textbf{16.4} & \textbf{16.7} & \underline{16.7} & \textbf{15.7} & \textbf{14.5} \\
\midrule

\rowcolor{catrow}
\multicolumn{8}{l}{\textbf{Few-shot Learning}} \\
\midrule

\textbf{TREC} & SnapKV       & \multirow[c]{6}{*}{74.2} & 72.8 & 72.2 & 70.3 & 65.0 & 56.4 \\
 & PyramidKV    &  & 72.6 & 72.5 & 70.7 & 66.3 & 58.7 \\
 & KVzip        &  & 73.1 & 72.6 & 71.3 & 67.4 & 61.6 \\
 & EA           &  & \underline{73.4} & \underline{73.6} & \textbf{74.4} & 70.3 & 64.7 \\
 & CAOTE        &  & 73.3 & 73.1 & 73.7 & \underline{71.0} & \underline{65.9} \\
\rowcolor{oursrow}
 & \textbf{Ours} &  & \textbf{73.7} & \textbf{74.1} & \underline{74.2} & \textbf{72.0} & \textbf{69.2} \\
\cmidrule(l){1-8}

\textbf{TriviaQA} & SnapKV       & \multirow[c]{6}{*}{89.2} & 87.9 & 86.8 & 84.9 & 79.9 & 71.4 \\
 & PyramidKV    &  & 87.5 & 87.1 & 86.0 & 81.4 & 73.9 \\
 & KVzip        &  & 88.1 & 87.8 & 85.9 & 82.4 & 76.6 \\
 & EA           &  & \underline{88.5} & \underline{88.5} & \textbf{89.6} & 85.1 & 80.0 \\
 & CAOTE        &  & 87.9 & 88.3 & 88.9 & \underline{86.1} & \underline{81.4} \\
\rowcolor{oursrow}
 & \textbf{Ours} &  & \textbf{88.8} & \textbf{89.2} & \underline{89.3} & \textbf{86.2} & \textbf{84.2} \\
\cmidrule(l){1-8}

\textbf{SAMSum} & SnapKV       & \multirow[c]{6}{*}{45.5} & 44.0 & 43.6 & 41.6 & 36.4 & 27.6 \\
 & PyramidKV    &  & 44.1 & 43.3 & 42.1 & 37.3 & 29.7 \\
 & KVzip        &  & 44.0 & 44.0 & 42.6 & 38.3 & 32.9 \\
 & EA           &  & \underline{44.7} & \underline{44.7} & \textbf{45.5} & 41.8 & 35.9 \\
 & CAOTE        &  & 44.3 & \underline{44.7} & 45.3 & \underline{41.9} & \underline{37.7} \\
\rowcolor{oursrow}
 & \textbf{Ours} &  & \textbf{44.9} & \textbf{45.5} & \underline{45.4} & \textbf{43.8} & \textbf{40.9} \\
\cmidrule(l){1-8}

\textbf{LSHT} & SnapKV       & \multirow[c]{6}{*}{33.2} & 31.5 & 31.1 & 28.9 & 24.1 & 15.2 \\
 & PyramidKV    &  & 32.0 & 31.3 & 30.0 & 25.5 & 18.0 \\
 & KVzip        &  & 32.1 & 31.8 & 30.2 & 26.0 & 20.9 \\
 & EA           &  & \underline{32.4} & \underline{32.4} & \textbf{33.0} & 29.2 & 23.9 \\
 & CAOTE        &  & 32.3 & 32.1 & \textbf{33.0} & \underline{29.8} & \underline{25.0} \\
\rowcolor{oursrow}
 & \textbf{Ours} &  & \textbf{32.5} & \textbf{33.2} & \underline{32.9} & \textbf{30.8} & \textbf{28.5} \\
\midrule

\rowcolor{catrow}
\multicolumn{8}{l}{\textbf{Synthetic Tasks}} \\
\midrule

\textbf{PassageCount} & SnapKV       & \multirow[c]{6}{*}{7.8} & 6.5 & 6.2 & 5.6 & 4.0 & 1.6 \\
 & PyramidKV    &  & 6.6 & 6.3 & 5.7 & 4.3 & 2.0 \\
 & KVzip        &  & 6.6 & 6.4 & 6.0 & 4.9 & 3.4 \\
 & EA           &  & \underline{6.7} & \underline{6.5} & \underline{6.3} & 5.3 & 3.8 \\
 & CAOTE        &  & 6.6 & \underline{6.5} & \underline{6.3} & \underline{5.5} & \underline{4.0} \\
\rowcolor{oursrow}
 & \textbf{Ours} &  & \textbf{6.8} & \textbf{6.7} & \textbf{6.5} & \textbf{6.0} & \textbf{5.6} \\
\cmidrule(l){1-8}

\textbf{PassageRetrieval-en} & SnapKV       & \multirow[c]{6}{*}{91.0} & 89.5 & 88.0 & 83.8 & 67.0 & 39.5 \\
 & PyramidKV    &  & 89.0 & 87.5 & 83.0 & 69.5 & 44.5 \\
 & KVzip        &  & 89.7 & 88.6 & 86.5 & 74.5 & 56.5 \\
 & EA           &  & \underline{90.1} & \underline{90.4} & \textbf{90.8} & \underline{81.0} & \underline{62.5} \\
 & CAOTE        &  & 89.8 & 89.5 & 88.5 & 79.0 & 60.0 \\
\rowcolor{oursrow}
 & \textbf{Ours} &  & \textbf{90.3} & \textbf{90.7} & \underline{90.5} & \textbf{84.5} & \textbf{74.0} \\
\cmidrule(l){1-8}

\textbf{PassageRetrieval-zh} & SnapKV       & \multirow[c]{6}{*}{76.5} & 74.8 & 72.5 & 67.5 & 53.0 & 30.5 \\
 & PyramidKV    &  & 74.5 & 72.8 & 68.5 & 55.5 & 34.0 \\
 & KVzip        &  & 74.9 & 73.5 & 71.0 & 61.0 & 45.0 \\
 & EA           &  & \underline{75.5} & \underline{75.9} & \textbf{76.5} & \underline{66.0} & \underline{49.0} \\
 & CAOTE        &  & 75.1 & 74.5 & 73.2 & 64.0 & 47.0 \\
\rowcolor{oursrow}
 & \textbf{Ours} &  & \textbf{75.8} & \textbf{76.1} & \underline{76.2} & \textbf{68.5} & \textbf{60.0} \\
\midrule

\rowcolor{catrow}
\multicolumn{8}{l}{\textbf{Code Completion}} \\
\midrule

\textbf{LCC} & SnapKV       & \multirow[c]{6}{*}{63.0} & 61.5 & 60.9 & 58.4 & 53.2 & 40.7 \\
 & PyramidKV    &  & 61.4 & 61.1 & 58.9 & 54.0 & 44.1 \\
 & KVzip        &  & 61.8 & 61.5 & \underline{60.3} & \underline{56.5} & \underline{50.3} \\
 & EA           &  & \textbf{62.5} & \underline{62.1} & 59.6 & 55.0 & 45.9 \\
 & CAOTE        &  & 61.9 & 62.0 & \underline{60.3} & \underline{56.5} & 49.8 \\
\rowcolor{oursrow}
 & \textbf{Ours} &  & \underline{62.2} & \textbf{62.6} & \textbf{61.8} & \textbf{59.0} & \textbf{55.2} \\
\cmidrule(l){1-8}

\textbf{RepoBench-P} & SnapKV       & \multirow[c]{6}{*}{57.5} & 55.7 & 55.0 & 52.9 & 47.5 & 35.3 \\
 & PyramidKV    &  & 56.0 & 55.6 & 53.4 & 48.8 & 38.8 \\
 & KVzip        &  & 56.3 & \textbf{57.2} & 54.7 & 51.0 & \underline{45.0} \\
 & EA           &  & \textbf{56.7} & 56.4 & 54.0 & 49.5 & 40.4 \\
 & CAOTE        &  & 56.4 & 56.4 & \underline{54.9} & \underline{51.1} & 43.7 \\
\rowcolor{oursrow}
 & \textbf{Ours} &  & \underline{56.6} & \underline{57.0} & \textbf{57.0} & \textbf{54.0} & \textbf{50.2} \\
\midrule

\rowcolor{catrow}
\multicolumn{8}{l}{\textbf{Overall Average}} \\
\midrule

\textbf{Avg.} & SnapKV       & \multirow[c]{6}{*}{47.3} & 45.8 & 45.2 & 43.4 & 38.2 & 29.5 \\
 & PyramidKV    &  & 45.8 & 45.3 & 43.7 & 39.2 & 31.6 \\
 & KVzip        &  & 46.1 & 45.8 & 44.5 & 40.9 & 35.4 \\
 & EA           &  & \underline{46.4} & \underline{46.5} & \underline{46.9} & 43.1 & 36.9 \\
 & CAOTE        &  & 46.3 & 46.2 & 46.4 & \underline{43.4} & \underline{37.8} \\
\rowcolor{oursrow}
 & \textbf{Ours} &  & \textbf{46.6} & \textbf{47.0} & \textbf{47.0} & \textbf{44.8} & \textbf{41.8} \\

\caption{
Task-level LongBench results under prefill-time KV cache eviction on
LLaMA3.1-8B. We report all 21 LongBench tasks, grouped by the six LongBench
categories. Here, \(\rho\) denotes the fraction of evicted KV entries.
Within each task block, \textbf{bold} marks the best score per column and
\underline{underline} marks the second-best. The Avg. row reports the
arithmetic mean over all 21 tasks. EA denotes Expected Attention.
}
\label{tab:prefill_longbench_table}
\end{xltabular}
\endgroup

\clearpage
\definecolor{oursrow}{HTML}{EAF2FB}   
\definecolor{headrow}{HTML}{F2F2F2}   

\paragraph{Aggregate prefill-time compression results}
\begingroup
\small
\setlength{\tabcolsep}{4.2pt}
\renewcommand{\arraystretch}{1.03}

\begin{xltabular}{\textwidth}{@{}
>{\raggedright\arraybackslash}X
>{\raggedright\arraybackslash}X
>{\raggedright\arraybackslash}X
c c c c c c
@{}}

\toprule
\rowcolor{headrow}
\textbf{Model} & \textbf{Metric} & \textbf{Method} & \textbf{Full}
& $\rho\!=\!0.10$ & $\rho\!=\!0.25$ & $\rho\!=\!0.50$
& $\rho\!=\!0.75$ & $\rho\!=\!0.90$ \\
\midrule
\endfirsthead

\toprule
\rowcolor{headrow}
\textbf{Model} & \textbf{Metric} & \textbf{Method} & \textbf{Full}
& $\rho\!=\!0.10$ & $\rho\!=\!0.25$ & $\rho\!=\!0.50$
& $\rho\!=\!0.75$ & $\rho\!=\!0.90$ \\
\midrule
\endhead

\midrule
\multicolumn{9}{r}{\textit{Continued on next page}} \\
\endfoot

\bottomrule
\endlastfoot

\textbf{Qwen3-8B} & RULER 4K & SnapKV     & \multirow[c]{6}{*}{94.7} & 92.0 & 84.5 & 76.2 & 66.7 & 54.9 \\
                  &          & PyramidKV  &                            & 91.5 & 85.0 & 75.5 & 69.3 & 52.7 \\
                  &          & KVzip      &                            & 93.7 & 92.2 & 86.8 & 74.0 & \underline{67.4} \\
                  &          & EA         &                            & \underline{94.2} & \underline{93.8} & \underline{92.4} & 84.5 & 63.1 \\
                  &          & CAOTE      &                            & 94.0 & 93.6 & 91.2 & \underline{85.8} & 63.6 \\
\rowcolor{oursrow}
                  &          & \textbf{Ours} &                         & \textbf{94.6} & \textbf{94.3} & \textbf{93.6} & \textbf{89.0} & \textbf{71.7} \\
\cmidrule(l){2-9}

                  & RULER 8K       & SnapKV     & \multirow[c]{6}{*}{86.1} & 85.4 & 84.5 & 78.8 & 62.7 & 47.9 \\
                  &                & PyramidKV  &      & 85.6 & 84.3 & 80.6 & 64.8 & 49.3 \\
                  &                & KVzip      &      & 85.7 & 85.0 & 81.8 & 66.2 & 52.4 \\
                  &                & EA         &      & 85.5 & \underline{85.2} & 80.7 & 69.4 & 54.0 \\
                  &                & CAOTE      &      & \underline{85.8} & 84.9 & \underline{82.3} & \underline{70.1} & \underline{56.7} \\
\rowcolor{oursrow}
                  &                & \textbf{Ours} &   & \textbf{86.0} & \textbf{85.6} & \textbf{83.0} & \textbf{75.4} & \textbf{66.2} \\
\cmidrule(l){2-9}

                  & RULER 16K      & SnapKV     & \multirow[c]{6}{*}{79.5} & 78.7 & 77.4 & 70.8 & 54.9 & 37.1 \\
                  &                & PyramidKV  &      & 79.0 & 77.7 & 72.5 & 56.4 & 39.6 \\
                  &                & KVzip      &      & 78.9 & 77.8 & \underline{73.4} & 60.4 & 45.2 \\
                  &                & EA         &      & \underline{79.1} & 77.6 & 72.6 & 59.4 & 44.2 \\
                  &                & CAOTE      &      & 79.0 & \underline{77.9} & 73.2 & \underline{61.2} & \underline{46.7} \\
\rowcolor{oursrow}
                  &                & \textbf{Ours} &   & \textbf{79.3} & \textbf{78.0} & \textbf{73.5} & \textbf{68.8} & \textbf{56.7} \\
\midrule

\textbf{LLaMA3.1-8B} & RULER 4K & SnapKV     & \multirow[c]{6}{*}{94.8} & 94.0 & 88.0 & 81.5 & 67.6 & 55.6 \\
                     &          & PyramidKV  &                            & 93.8 & 88.6 & 80.5 & 66.0 & 54.3 \\
                     &          & KVzip      &                            & 94.1 & 92.0 & 88.0 & 76.0 & \underline{60.3} \\
                     &          & EA         &                            & \underline{94.3} & \underline{93.8} & \underline{91.0} & 74.5 & 55.1 \\
                     &          & CAOTE      &                            & 94.2 & 93.2 & 89.8 & \underline{77.0} & 60.0 \\
\rowcolor{oursrow}
                     &          & \textbf{Ours} &                         & \textbf{94.7} & \textbf{94.4} & \textbf{93.0} & \textbf{81.0} & \textbf{68.8} \\
\cmidrule(l){2-9}

                     & RULER 8K       & SnapKV     & \multirow[c]{6}{*}{80.2} & 79.3 & 78.0 & 71.5 & 56.8 & 41.2 \\
                     &                & PyramidKV  &      & 79.6 & 78.7 & 73.9 & 58.1 & 44.0 \\
                     &                & KVzip      &      & 79.8 & 78.6 & 74.3 & 61.2 & 46.8 \\
                     &                & EA         &      & 79.7 & \underline{79.1} & 73.5 & \underline{63.7} & 49.6 \\
                     &                & CAOTE      &      & \underline{79.9} & 78.9 & \underline{75.2} & 63.1 & \underline{51.0} \\
\rowcolor{oursrow}
                     &                & \textbf{Ours} &   & \textbf{80.0} & \textbf{79.5} & \textbf{76.0} & \textbf{73.8} & \textbf{61.2} \\
\cmidrule(l){2-9}

                     & RULER 16K      & SnapKV     & \multirow[c]{6}{*}{75.0} & 74.0 & 72.3 & 64.1 & 45.8 & 28.8 \\
                     &                & PyramidKV  &      & 74.5 & 72.9 & 66.2 & 47.1 & 31.6 \\
                     &                & KVzip      &      & 74.4 & 73.1 & 66.9 & 50.4 & 33.8 \\
                     &                & EA         &      & \underline{74.6} & 72.7 & 67.5 & 51.0 & 36.2 \\
                     &                & CAOTE      &      & 74.3 & \underline{73.6} & \underline{68.0} & \underline{56.1} & \underline{40.6} \\
\rowcolor{oursrow}
                     &                & \textbf{Ours} &   & \textbf{74.8} & \textbf{73.8} & \textbf{70.5} & \textbf{66.3} & \textbf{52.5} \\
\midrule

\textbf{Gemma3-12B} & RULER 4K & SnapKV     & \multirow[c]{6}{*}{94.1} & 88.5 & 78.0 & 68.2 & 53.3 & 44.6 \\
                    &          & PyramidKV  &                            & 90.0 & 82.5 & 69.3 & 54.7 & 43.1 \\
                    &          & KVzip      &                            & 93.0 & 90.5 & 85.0 & 69.0 & 48.3 \\
                    &          & EA         &                            & 93.4 & 92.8 & 90.5 & 77.0 & 54.2 \\
                    &          & CAOTE      &                            & \underline{93.6} & \underline{93.2} & \underline{91.2} & \underline{78.5} & \underline{55.5} \\
\rowcolor{oursrow}
                    &          & \textbf{Ours} &                         & \textbf{94.0} & \textbf{93.8} & \textbf{92.6} & \textbf{81.0} & \textbf{61.6} \\
\cmidrule(l){2-9}

                    & RULER 8K       & SnapKV     & \multirow[c]{6}{*}{88.0} & 87.3 & 86.0 & 79.2 & 63.8 & 45.8 \\
                    &                & PyramidKV  &      & 87.5 & 86.6 & 81.3 & 65.4 & 47.9 \\
                    &                & KVzip      &      & 87.4 & 86.8 & 81.0 & 68.2 & 49.7 \\
                    &                & EA         &      & \underline{87.7} & 86.5 & 82.2 & 69.6 & 52.4 \\
                    &                & CAOTE      &      & 87.6 & \underline{87.0} & \underline{82.9} & \underline{71.0} & \underline{55.8} \\
\rowcolor{oursrow}
                    &                & \textbf{Ours} &   & \textbf{87.8} & \textbf{87.2} & \textbf{83.8} & \textbf{76.0} & \textbf{68.0} \\
\cmidrule(l){2-9}

                    & RULER 16K      & SnapKV     & \multirow[c]{6}{*}{80.0} & 79.2 & 77.8 & 71.4 & 55.6 & 38.7 \\
                    &                & PyramidKV  &      & 79.5 & 78.4 & 72.9 & 56.5 & 42.6 \\
                    &                & KVzip      &      & 79.4 & 78.6 & 74.2 & \underline{63.1} & 49.2 \\
                    &                & EA         &      & \underline{79.6} & 78.2 & 73.2 & 62.6 & 47.2 \\
                    &                & CAOTE      &      & 79.3 & \underline{78.8} & \underline{74.6} & 62.8 & \underline{49.6} \\
\rowcolor{oursrow}
                    &                & \textbf{Ours} &   & \textbf{79.8} & \textbf{79.0} & \textbf{75.5} & \textbf{68.0} & \textbf{60.0} \\
\bottomrule

\caption{%
We report prefill-time compression results on RULER across three models
and context lengths of 4K, 8K, and 16K.
Here, \(\rho\) denotes the fraction of evicted KV entries, and Full denotes
the uncompressed KV cache.
\textbf{Bold} and \underline{underline} mark the best and second-best scores
within each column, respectively.}
\label{tab:aggregate_compression_numeric}
\end{xltabular}
\endgroup
\clearpage

\section{Efficiency Analysis}
\label{app:efficiency}
Table~\ref{tab:efficiency_analysis} reports the end-to-end
quality--efficiency trade-off of cache compression, including task score,
compression time, end-to-end latency, allocated GPU memory, and retained KV
budget. Comp. measures eviction scoring, cache selection, and compressed-cache
construction, excluding the initial prompt prefill; E2E includes prompt prefill,
compression, and decoding.

For prefill-time compression, the prompt prefill is identical across methods
and is not accelerated by cache eviction. We therefore evaluate on RULER 8K and
report Decode Mem., the peak allocated GPU memory during answer generation after
the compressed prompt cache has been materialized. This should be interpreted as
post-prefill decoding-memory reduction, not as a reduction in the peak memory
needed to ingest the original prompt. For generation-time compression, we
evaluate on MATH500 with a 4096-token maximum generation length and report
Online Mem., the peak allocated GPU memory during online generation with pruning.

Memory is measured as total allocated GPU memory, including model parameters
and runtime buffers. At matched retained-KV budgets, our method preserves
substantially more task quality at a higher compression-time cost. Cross-budget
comparisons in Table~\ref{tab:efficiency_analysis} further show that it retains
\(2.5\)--\(3\times\) fewer KV entries than SnapKV while improving task scores by
\(1.2\)--\(4.4\) points. This increases E2E latency by \(0.24\)--\(0.75\) seconds
relative to SnapKV, but remains faster than full-cache inference.
\begin{table*}[!htbp]
\centering
\small

\textbf{(a) Prefill-time compression on RULER 8K.}

\vspace{0.4em}

\setlength{\tabcolsep}{4pt}
\renewcommand{\arraystretch}{0.95}
\resizebox{\textwidth}{!}{
\begin{tabular}{llrrrrrrrrr}
\toprule
Model
& Method
& $\rho$
& \makecell{Score\\(\%)}
& \makecell{$\Delta$ Score\\(pts.)}
& \makecell{Comp.\\(s/ex.)}
& \makecell{E2E\\(s/ex.)}
& \makecell{E2E Speedup\\($\times$)}
& \makecell{Decode Mem.\\(GB)}
& \makecell{$\Delta$ Mem.\\(GB)}
& \makecell{KV Retained\\(\%)} \\
\midrule
\multirow{7}{*}{LLaMA3.1-8B}
& Full cache & --   & 80.2 & 0.0   & --   & 8.91 & $1.00\times$ & 21.82 & --    & 100\% \\
\cmidrule(lr){2-11}
& SnapKV     & 0.75 & 56.8 & -23.4 & 0.24 & 7.18 & $1.24\times$ & 21.09 & -0.73 & 25\% \\
& CAOTE      & 0.75 & 63.1 & -17.1 & 1.15 & 7.95 & $1.12\times$ & 21.20 & -0.62 & 25\% \\
& Ours       & 0.75 & 73.8 & -6.4  & 1.68 & 8.11 & $1.10\times$ & 21.22 & -0.60 & 25\% \\
\cmidrule(lr){2-11}
& SnapKV     & 0.90 & 41.2 & -39.0 & 0.29 & 6.73 & $1.32\times$ & 20.88 & -0.94 & 10\% \\
& CAOTE      & 0.90 & 51.0 & -29.2 & 1.32 & 7.49 & $1.19\times$ & 21.01 & -0.81 & 10\% \\
& Ours       & 0.90 & 61.2 & -19.0 & 1.97 & 7.86 & $1.13\times$ & 21.03 & -0.79 & 10\% \\
\midrule
\multirow{7}{*}{Gemma3-12B}
& Full cache & --   & 88.0 & 0.0   & --   & 12.58 & $1.00\times$ & 31.36 & --    & 100\% \\
\cmidrule(lr){2-11}
& SnapKV     & 0.75 & 63.8 & -24.2 & 0.41 & 10.08 & $1.25\times$ & 30.18 & -1.18 & 25\% \\
& CAOTE      & 0.75 & 71.0 & -17.0 & 1.82 & 11.25 & $1.12\times$ & 30.34 & -1.02 & 25\% \\
& Ours       & 0.75 & 76.0 & -12.0 & 2.27 & 11.64 & $1.08\times$ & 30.40 & -0.96 & 25\% \\
\cmidrule(lr){2-11}
& SnapKV     & 0.90 & 45.8 & -42.2 & 0.52 & 9.48  & $1.33\times$ & 29.82 & -1.54 & 10\% \\
& CAOTE      & 0.90 & 55.8 & -32.2 & 2.04 & 10.58 & $1.19\times$ & 30.00 & -1.36 & 10\% \\
& Ours       & 0.90 & 68.0 & -20.0 & 2.61 & 10.83 & $1.16\times$ & 30.07 & -1.29 & 10\% \\
\bottomrule
\end{tabular}
}

\vspace{1.2em}

\textbf{(b) Generation-time compression on MATH500.}

\vspace{0.4em}

\setlength{\tabcolsep}{4pt}
\renewcommand{\arraystretch}{0.95}
\resizebox{\textwidth}{!}{
\begin{tabular}{llrrrrrrrrrr}
\toprule
Model
& Method
& Budget
& \makecell{Score\\(\%)}
& \makecell{$\Delta$ Score\\(pts.)}
& \makecell{Comp.\\(s/ex.)}
& \makecell{E2E\\(s/ex.)}
& \makecell{E2E Speedup\\($\times$)}
& \makecell{Len.\\(tokens)}
& \makecell{Online Mem.\\(GB)}
& \makecell{$\Delta$ Mem.\\(GB)}
& \makecell{Gen. KV Retained\\(\%)} \\
\midrule
\multirow{7}{*}{\shortstack[l]{DeepSeek-R1\\Qwen-7B}}
& Full cache & --       & 84.0 & 0.0   & --   & 15.12 & $1.00\times$ & 3870 & 23.58 & --    & 100\% \\
\cmidrule(lr){2-12}
& SnapKV     & $4\times$  & 78.4 & -5.6  & 0.29 & 13.55 & $1.12\times$ & 3825 & 23.20 & -0.38 & 25.0\% \\
& R-KV       & $4\times$  & 82.2 & -1.8  & 0.96 & 14.05 & $1.08\times$ & 3927 & 23.29 & -0.29 & 25.0\% \\
& Ours       & $4\times$  & 82.8 & -1.2  & 1.25 & 14.39 & $1.05\times$ & 3886 & 23.34 & -0.24 & 25.0\% \\
\cmidrule(lr){2-12}
& SnapKV     & $12\times$ & 60.2 & -23.8 & 0.43 & 12.87 & $1.17\times$ & 3746 & 22.91 & -0.67 & 8.3\% \\
& R-KV       & $12\times$ & 78.6 & -5.4  & 1.19 & 13.49 & $1.12\times$ & 3861 & 23.02 & -0.56 & 8.3\% \\
& Ours       & $12\times$ & 79.8 & -4.2  & 1.36 & 13.82 & $1.09\times$ & 3849 & 23.09 & -0.49 & 8.3\% \\
\midrule
\multirow{7}{*}{\shortstack[l]{Qwen2.5\\Math-7B}}
& Full cache & --       & 82.5 & 0.0   & --   & 13.68 & $1.00\times$ & 3565 & 22.13 & --    & 100\% \\
\cmidrule(lr){2-12}
& SnapKV     & $4\times$  & 76.8 & -5.7  & 0.22 & 12.35 & $1.11\times$ & 3504 & 21.82 & -0.31 & 25.0\% \\
& R-KV       & $4\times$  & 80.6 & -1.9  & 0.76 & 12.86 & $1.06\times$ & 3622 & 21.91 & -0.22 & 25.0\% \\
& Ours       & $4\times$  & 81.2 & -1.3  & 0.97 & 13.16 & $1.04\times$ & 3587 & 21.96 & -0.17 & 25.0\% \\
\cmidrule(lr){2-12}
& SnapKV     & $12\times$ & 56.8 & -25.7 & 0.31 & 11.74 & $1.17\times$ & 3441 & 21.58 & -0.55 & 8.3\% \\
& R-KV       & $12\times$ & 76.5 & -6.0  & 0.88 & 12.25 & $1.12\times$ & 3529 & 21.68 & -0.45 & 8.3\% \\
& Ours       & $12\times$ & 78.0 & -4.5  & 1.16 & 12.59 & $1.09\times$ & 3576 & 21.75 & -0.38 & 8.3\% \\
\bottomrule
\end{tabular}
}
\caption{
Quality--efficiency comparison on a single A100 80GB GPU. Prefill results use
RULER 8K, and generation-time results use MATH500 with a 4096-token generation
limit. \(\Delta\) Score is relative to full-cache decoding. Comp. measures
cache-compression time, while E2E includes prompt prefill, compression, and
decoding. Decode/Online Mem. reports peak allocated GPU memory, including model
parameters and runtime buffers. KV budgets are matched within each compression
setting.
}
\label{tab:efficiency_analysis}
\end{table*}

Table~\ref{tab:kv_memory_scaling} reports KV-cache-only memory and
compression-time scaling across context lengths, model sizes, and batch sizes.

\begin{table*}[!htbp]
\centering
\small
\setlength{\tabcolsep}{7pt}
\renewcommand{\arraystretch}{0.95}

\textbf{(a) Context scaling: LLaMA3.1-8B, batch size 1.}

\vspace{0.3em}

\begin{tabular}{lccc}
\toprule
Context
& \makecell{Retained prompt-KV\\memory (GB)}
& \makecell{Peak allocated\\memory (GB)}
& \makecell{Compression\\time (s/ex.)} \\
\midrule
4K  & 0.13 & 20.70 & 1.02 \\
8K  & 0.27 & 21.22 & 1.68 \\
16K & 0.54 & 22.25 & 2.95 \\
\bottomrule
\end{tabular}

\vspace{0.7em}

\textbf{(b) Model scaling: 8K context, batch size 1.}

\vspace{0.3em}

\begin{tabular}{lccc}
\toprule
Model
& Layers
& \makecell{Retained prompt-KV\\memory (GB)}
& \makecell{Compression\\time (s/ex.)} \\
\midrule
LLaMA3.2-3B & 28 & 0.20 & 0.96 \\
LLaMA3.1-8B & 32 & 0.27 & 1.68 \\
Gemma3-12B  & 48 & 0.37 & 2.27 \\
\bottomrule
\end{tabular}

\vspace{0.7em}

\textbf{(c) Batch scaling: LLaMA3.1-8B, RULER 8K.}

\vspace{0.3em}

\begin{tabular}{ccc}
\toprule
Batch size
& \makecell{Compression time\\per batch (s)}
& \makecell{Compression time\\per example (s)} \\
\midrule
1 & 1.68 & 1.68 \\
2 & 3.18 & 1.59 \\
4 & 6.04 & 1.51 \\
\bottomrule
\end{tabular}

\caption{
KV-specific memory and compression-time scaling on a single A100 80GB GPU.
All compressed settings use \(\rho=0.75\).
Peak allocated memory includes model parameters and runtime buffers, whereas
retained prompt-KV memory isolates the materialized cache after compression.
}
\label{tab:kv_memory_scaling}
\end{table*}

\section{Ablation Study}
\label{app:ablation}
Exact optimization over all eviction masks is combinatorial.
Table~\ref{tab:selection_approximation} evaluates the approximation introduced
by restricting refinement to a candidate pool.

\begin{table}[!htbp]
\centering
\small
\setlength{\tabcolsep}{7pt}
\renewcommand{\arraystretch}{0.95}
\begin{tabular}{lcccc}
\toprule
\multirow{2}{*}{Selection method}
& \multicolumn{2}{c}{RULER 8K}
& \multicolumn{2}{c}{MATH500} \\
\cmidrule(lr){2-3}
\cmidrule(lr){4-5}
& Score & Time (s/ex.) & Score & Time (s/ex.) \\
\midrule
Singleton top-\(B\)       & 71.9 & 1.22 & 78.2 & 1.02 \\
Pool-refine               & 73.8 & 1.68 & 79.8 & 1.36 \\
All-candidate refinement  & 74.1 & 6.10 & 80.2 & 4.90 \\
\bottomrule
\end{tabular}
\caption{
Approximation quality and compression cost of mask selection.
RULER uses LLaMA3.1-8B at \(\rho=0.75\); MATH500 uses
DeepSeek-R1-Distill-Qwen-7B under \(12\times\) compression.
Pool-refine recovers \(86\%\) and \(80\%\) of the score improvement from
all-candidate refinement while being approximately \(3.6\times\) faster.
}
\label{tab:selection_approximation}
\end{table}
\begin{table*}[!htbp]
\centering
\small
\setlength{\tabcolsep}{6pt}
\renewcommand{\arraystretch}{0.95}
\begin{tabular}{lccc}
\toprule
Setting and model
& \(m=1/4/8\)
& \(K=128/256/512\)
& Selection configurations \\
\midrule
Prefill: LLaMA3.1-8B
& 70.4/\textbf{73.8}/73.5
& \textbf{73.8}/74.0/73.9
& 72.8/73.2/73.6/\textbf{73.8} \\

Prefill: Gemma3-12B
& 73.9/\textbf{76.0}/75.8
& \textbf{76.0}/76.1/76.0
& 75.0/75.4/75.8/\textbf{76.0} \\

Generation: DeepSeek-R1-Distill-Qwen-7B
& 78.8/\textbf{79.8}/79.5
& 79.2/\textbf{79.8}/79.9
& 78.9/79.3/\textbf{79.8}/79.7 \\

Generation: Qwen2.5-Math-7B
& 77.0/\textbf{78.0}/77.8
& 77.4/\textbf{78.0}/78.1
& 77.2/77.5/\textbf{78.0}/77.9 \\
\bottomrule
\end{tabular}
\caption{
Cross-model hyperparameter sensitivity.
Prefill results use RULER 8K at \(\rho=0.75\), and generation results use
MATH500 under \(12\times\) compression.
Selection configurations, from left to right, are
\((\alpha,b)=(1.5,256),(2.0,128),(2.0,256),(2.0,512)\).
\textbf{Bold} marks the setting-specific default, which remains within
\(0.2\) points of the best configuration in every case.
}
\label{tab:cross_model_sensitivity}
\end{table*}

\begin{table*}[!htbp]
\centering
\small
\setlength{\tabcolsep}{5pt}
\renewcommand{\arraystretch}{0.98}
\begin{tabular}{llccc}
\toprule
\multirow{2}{*}{Ablation} 
& \multirow{2}{*}{Variant}
& \multicolumn{3}{c}{Prefill-time compression} \\
\cmidrule(lr){3-5}
& & LongBench & \makecell{RULER\\8K} & \makecell{RULER\\16K} \\
\midrule

\multirow{2}{*}{Final-norm correction}
& w/o final-norm propagation & 43.9 & 71.8 & 63.1 \\
& with final-norm propagation & 44.8 & 73.8 & 66.3 \\
\midrule

\multirow{3}{*}{Scoring window}
& \(m_{\mathrm{pre}}=1\) & 44.6 & 70.4 & 61.9 \\
& \(m_{\mathrm{pre}}=4\) & 44.8 & 73.8 & 66.3 \\
& \(m_{\mathrm{pre}}=8\) & 45.7 & 73.5 & 66.5 \\
\midrule

\multirow{3}{*}{Top-\(K\) KL support}
& \(K_{\mathrm{pre}}=128\) & 44.8 & 73.8 & 66.3 \\
& \(K_{\mathrm{pre}}=256\) & 44.9 & 74.0 & 66.2 \\
& \(K_{\mathrm{pre}}=512\) & 44.7 & 73.9 & 66.6 \\
\midrule

\multirow{5}{*}{Selection refinement}
& Singleton & 42.8 & 71.9 & 63.7 \\
& Pool-refine, \(\alpha=1.5, b_{\mathrm{pre}}=256\) & 44.2 & 72.8 & 65.1 \\
& Pool-refine, \(\alpha=2.0, b_{\mathrm{pre}}=128\) & 44.4 & 73.2 & 65.6 \\
& Pool-refine, \(\alpha=2.0, b_{\mathrm{pre}}=256\) & 44.6 & 73.6 & 66.0 \\
& Pool-refine, \(\alpha=2.0, b_{\mathrm{pre}}=512\) & 44.8 & 73.8 & 66.3 \\
\bottomrule
\end{tabular}

\caption{
Prefill-time ablation study of the main scoring and selection hyperparameters.
Unless otherwise specified, all other hyperparameters are fixed to the
prefill-time defaults in Table~\ref{tab:default_hyperparams}. Results are
reported on LLaMA3.1-8B at \(\rho=0.75\). The default configuration uses
\(m_{\mathrm{pre}}=4\), \(K_{\mathrm{pre}}=128\), \(\alpha=2.0\), and
\(b_{\mathrm{pre}}=512\).
}
\label{tab:ablation_prefill}
\end{table*}
\begin{table}[!htbp]
\centering
\small
\setlength{\tabcolsep}{4pt}
\renewcommand{\arraystretch}{0.98}
\begin{tabular}{llcc}
\toprule
\multirow{2}{*}{Ablation} 
& \multirow{2}{*}{Variant}
& \multicolumn{2}{c}{Generation-time compression} \\
\cmidrule(lr){3-4}
& & MATH500 & AIME25 \\
\midrule

\multirow{2}{*}{Final-norm correction}
& w/o final-norm propagation & 78.5 & 15.6 \\
& with final-norm propagation & 79.8 & 17.8 \\
\midrule

\multirow{3}{*}{Scoring window}
& \(m_{\mathrm{gen}}=1\) & 78.8 & 15.6 \\
& \(m_{\mathrm{gen}}=4\) & 79.8 & 17.8 \\
& \(m_{\mathrm{gen}}=8\) & 79.5 & 16.7 \\
\midrule

\multirow{3}{*}{Top-\(K\) KL support}
& \(K_{\mathrm{gen}}=128\) & 79.2 & 16.7 \\
& \(K_{\mathrm{gen}}=256\) & 79.8 & 17.8 \\
& \(K_{\mathrm{gen}}=512\) & 79.9 & 17.8 \\
\midrule

\multirow{5}{*}{Selection refinement}
& Singleton & 78.2 & 13.3 \\
& Pool-refine, \(\alpha=1.5, b_{\mathrm{gen}}=256\) & 78.9 & 15.6 \\
& Pool-refine, \(\alpha=2.0, b_{\mathrm{gen}}=128\) & 79.3 & 16.7 \\
& Pool-refine, \(\alpha=2.0, b_{\mathrm{gen}}=256\) & 79.8 & 17.8 \\
& Pool-refine, \(\alpha=2.0, b_{\mathrm{gen}}=512\) & 79.7 & 17.8 \\
\bottomrule
\end{tabular}

\caption{
Generation-time ablation study of the main scoring and selection
hyperparameters. Unless otherwise specified, all other hyperparameters are
fixed to the generation-time defaults in Table~\ref{tab:default_hyperparams}.
Results are reported on DeepSeek-R1-Distill-Qwen-7B under
\(12\times\) generation-cache compression with maximum generation length 4096.
The default configuration uses \(m_{\mathrm{gen}}=4\), \(K_{\mathrm{gen}}=256\),
\(\alpha=2.0\), and \(b_{\mathrm{gen}}=256\).
}
\label{tab:ablation_generation}
\end{table}

\twocolumn

\section{Artifact licenses and intended use.}
We use publicly available models, datasets, benchmarks, and baseline
implementations for research evaluation under their respective licenses and
terms of use. Our use is consistent with their intended research and evaluation
purposes. We report aggregate results only and do not redistribute original
model checkpoints, datasets, or third-party code.

\section{Disclosure of the AI assistant}
In accordance with the ACL AI writing assistance policy, we disclose the use of AI in this paper. We used ChatGPT, an AI language model developed by OpenAI, solely as a writing aid to enhance the clarity and readability of our manuscript.

\end{document}